\documentclass{article}

\PassOptionsToPackage{table}{xcolor}
\usepackage{etoolbox}       %
\usepackage[utf8]{inputenc} 

\usepackage{graphicx}  %

\usepackage[parfill]{parskip}
\usepackage{amsmath,amsthm}
\usepackage{mathtools}  %

\newtoggle{arxiv}
\toggletrue{arxiv}

  \usepackage[parfill]{parskip}
  \usepackage[citestyle=authoryear-comp,sorting=nyt,maxbibnames=99, maxcitenames=1, backend=biber, uniquename=false]{biblatex}
  \newcommand{\citep}{\parencite}
  
  \newcommand{\citet}{\textcite}
  \newlength{\defbaselineskip}
  \RequirePackage[T1]{fontenc}
  \RequirePackage[tt=false, type1=true]{libertine}
  \RequirePackage[varqu]{zi4}
  \RequirePackage[libertine]{newtxmath}

\newcommand{\framework}[0]{\textsc{Miras}}
\newcommand{\ymodel}[0]{\textsc{Yaad}}
\newcommand{\mmodel}[0]{\textsc{Moneta}}
\newcommand{\xmodel}[0]{\textsc{Memora}}

\newcommand{\R}[0]{\mathbb{R}}
\newcommand{\vk}[0]{\mathbf{k}}
\newcommand{\vvv}[0]{\mathbf{v}}
\newcommand{\cW}[0]{\mathcal{W}}
\newcommand{\vq}[0]{\mathbf{q}}
\newcommand{\mb}[1]{\mathbf{#1}}

\newcommand{\inner}[2]{\langle #1, #2 \rangle}
\newcommand{\undermath}[2]{\underset{#1}{\underbrace{#2}}}

\newcommand{\M}[0]{\mathcal{M}}

\usepackage{bm}

\usepackage[inline]{enumitem}
\usepackage{booktabs}
\usepackage{subcaption}
\usepackage{multirow}
\usepackage{wrapfig}
\usepackage{algorithm,algorithmicx,algpseudocode}
\usepackage{nicematrix}

\usepackage[frozencache,cachedir=.]{minted} %

\usepackage{import}

\newtheorem{theorem}{Theorem}[section]  %

\newtheorem{remark}{Remark}

\usepackage{pifont}%
\usepackage[T1]{fontenc}    
\usepackage{url}            
\usepackage{booktabs}       
\usepackage{amsfonts}       
\usepackage{nicefrac}       
\usepackage{microtype}      

\usepackage{wrapfig}

\usepackage{caption}
\usepackage{capt-of}
\usepackage{lipsum}

\usepackage[colorlinks]{hyperref}
\usepackage[capitalise,noabbrev]{cleveref}

\usepackage{subcaption}
\usepackage{threeparttable}

\definecolor{darkblue}{rgb}{0, 0, 0.5}
\hypersetup{colorlinks=true, citecolor=darkblue, linkcolor=darkblue, urlcolor=darkblue}

\definecolor{c1}{HTML}{586770}
\definecolor{c4}{HTML}{2a4a67}
\definecolor{c3}{HTML}{6d2a58}
\definecolor{c2}{HTML}{34142a}
\hypersetup{colorlinks=true, citecolor=c3, linkcolor=c2, urlcolor=c2}

\definecolor{myblue}{HTML}{FDF5E0} 
\definecolor{mygray}{HTML}{DBE2E9} 
\definecolor{mygreen}{HTML}{E6F3FC}

\usepackage{multirow}

\theoremstyle{plain}
\newtheorem{proposition}[theorem]{Proposition}
\theoremstyle{definition}
\newtheorem{definition}[theorem]{Definition}

\theoremstyle{remark}

\newcommand{\head}[1]{\vspace{1.7mm}\noindent{{\textcolor{c2}{\bf #1.}}}}

\usepackage{mathtools}

\usepackage{tikz}

\definecolor{dark2orange}{rgb}{0.9, 0.4, 0.}
\definecolor{dark2purple}{rgb}{0.4, 0.4, 0.8}

\usepackage{authblk}

\title{\textcolor{c2}{It's All Connected}: A Journey Through Test-Time Memorization, Attentional Bias, Retention, and Online Optimization}

\author[$^\dagger$]{Ali Behrouz}
  \author[$^\dagger$]{Meisam Razaviyayn}
  \author[$^\dagger$]{Peilin Zhong}
  \author[$^\dagger$]{Vahab Mirrokni}
  \affil[$^\dagger$]{Google Research}
   \affil[ ]{{\small \texttt{\{alibehrouz, razaviyayn, peilinz, mirrokni\}@google.com}}}
  \date{}

\begin{document}

\maketitle

\begin{abstract}
Designing efficient and effective architectural backbones has been in the core of research efforts to enhance the capability of foundation models. Inspired by the human cognitive phenomenon of attentional bias—the natural tendency to prioritize certain events or stimuli—we reconceptualize neural architectures, including Transformers, Titans, and modern linear recurrent neural networks as associative memory modules that learn a mapping of keys and values using an internal objective, referred to as \emph{attentional bias}. Surprisingly, we observed that most existing sequence models leverage either (1) dot-product similarity, or (2) $\ell_2$ regression objectives as their attentional bias. Going beyond these  objectives, we present a set of alternative attentional bias configurations along with their effective approximations to stabilize their training procedure. We then reinterpret forgetting mechanisms in modern deep learning architectures as a form of \emph{retention} regularization, providing a novel set of forget gates for sequence models. Building upon these insights, we present \framework{}, a general framework to design deep learning architectures based on four choices of: (i) associative memory architecture, (ii) attentional bias objective, (iii) retention gate, and (iv) memory learning algorithm. We present three novel sequence models—\mmodel, \ymodel, and \xmodel—that go beyond the power of existing linear RNNs while maintaining a fast parallelizable training process. Our experiments show different design choices in \framework{} yield models with varying strengths. 
 For example, certain  instances of \framework{} achieve exceptional performance in special tasks such as language modeling, commonsense reasoning, and recall intensive tasks, even outperforming Transformers and other modern linear recurrent models. 
\end{abstract}

\section{Introduction}\label{sec:intro}
Designing efficient architectural backbones for sequence modeling is a key to enhance the capability of foundation models in domains ranging from language~\citep{Vaswani+2017, behrouz2024titans} and computer vision~\citep{dosovitskiy2020image} to computational biology~\citep{wang2024long} and neuroscience~\citep{behrouz2024unsupervised}. While Transformers~\citep{Vaswani+2017}, mainly due to their in-context learning and ability to learn at scale~\citep{kaplan2020scaling}, have been firmly established as state-of-the-art (SOTA) models in sequence modeling, their quadratic time and space complexity limits their applicability in tasks that require long context modeling~\citep{dalal2025one, liu2024lost, li2024survey}. 

Recent efforts aim to overcome Transformer limitations in long-context modeling by designing efficient recurrent alternatives~\citep{neil2017delta, smith2022simplified, behrouz2024titans}. Unlike Transformer's linearly growing memory (i.e., the KV cache), these models compress the context into a fixed size memory, demanding improved memory management for comparable performance. To design more effective architectures, studies focus on improving memory capacity and its management by using/designing more expressive: (1) Learning rules: from Hebbian rule~\citep{hebb2005organization} to Delta rule~\citep{neil2017delta}; (2) Forget gates: from LSTM's~\citep{LSTM} to Mamba2's~\citep{dao2024transformers} and then Titan's forget gates~\citep{behrouz2024titans}; and (3) More expressive memory architectures: from vector-valued memory in RetNet~\citep{sun2023retentive} and LRU~\citep{orvieto2023resurrecting} to neural deep memory in Titans~\citep{behrouz2024titans} and TTT~\citep{sun2024learning}.

At the core of these advancements lies a critical question: ``what is the underlying design framework behind these sequence models, and how can these models be enhanced?''. Taking inspiration from the broad definitions of associative memory and learning in neuropsychology literature~\citep{okano2000learning}, several studies discuss the connection between Transformers and (linear) Recurrent Neural Networks (RNNs) with associative memory~\citep{hopfield1982neural, ramsauer2021hopfield, bietti2023birth}. These studies, however, either: (1) lack a \emph{universal} explanation to \emph{fully} illustrate the underlying learning algorithms, (2) are limited to a specific definition of associative memory and lack generalizability, and/or (3) are unable to describe standard, widely used components such as forget gate.

\head{Contributions}
Inspired by the human cognitive phenomenon of attentional bias—the natural tendency to prioritize certain events or stimuli—we re-conceptualize neural architectures, including Transformers, Titans, and other modern linear recurrent neural networks based on a \emph{broad} definition of associative memory with \emph{attentional bias}. We define and formalize the concept of {attentional bias} as the internal memory objective of sequence models (see \autoref{sec:AM}) that aims to learn the underlying mapping between inputs (i.e., keys and values). Our formulation reveals that almost all existing sequence models are associative memories that leverage the same type of attentional bias. We reinterpret existing forgetting mechanisms in modern deep learning architectures as a form of \emph{retention} $\ell_2$-regularization for the attentional bias, and then provide a novel set of alternative retention gates (forget gate) for sequence models, providing new insights on how to balance learning new concepts and the retention of previously learned concepts.

Building upon our formulation of memory and forget gate, we present \framework\footnote{\: ``Miras'' is the translation of ``Legacy'' in several languages: such as Persian, Arabic, and Turkish. We choose this name since this framework provides clear steps for future studies to design powerful sequence models based on their task at hand.}, a fundamental framework to design novel sequence modeling architectures by four choice of: (1) Attentional bias (i.e., memory objective), (2) Retention gate, (3) Memory architecture, and (4) Memory learning algorithm (i.e., optimizer). We motivate and discuss several novel design choices, leading to novel architectures beyond existing sequence modeling architectures.

Finally, we focus on three novel variants of \framework—\mmodel
, \ymodel
, and \xmodel
—that are based on attentional biases beyond simple $\ell_2$-regression objective as well as novel retention gating mechanisms that are more robust than existing ones. We further perform experimental evaluations of these three variants on language modeling, common-sense reasoning, needle-in-haystack, and recall intensive tasks. The results illustrates the superior performance of these variants, outperforming state-of-the-art sequence models. 

\head{Roadmap} In \autoref{sec:background}, we review literature and discuss relevant concepts that we use through the paper. In \autoref{sec:AM}, we present and discuss the broad definition of associative memory with formally defining the concept of \emph{attentional bias}. We then discuss two viewpoints—Learning-Retaining and Follow-the-Regularized-Leader (FTRL)—to interpret sequence modeling through the lens of optimization and prove the generality of Learning-Retaining over FTRL. In \autoref{sec:miras}, we present our \framework{} framework and discuss how it unifies modern sequence models. In \autoref{sec:beyond-variants}, to show the potential of \framework{} framework, we discuss a variety of novel design choices for (1) attentional bias, and (2) retention gate (forget gate). Later in \autoref{sec:Miras-variants}, we present three novel sequence models as the variants of \framework, and then discuss how to train them in a parallelizable manner. Finally, our experimental evaluations are reported in \autoref{sec:exp}.

\begin{figure*}
    \centering
    \includegraphics[width=\linewidth]{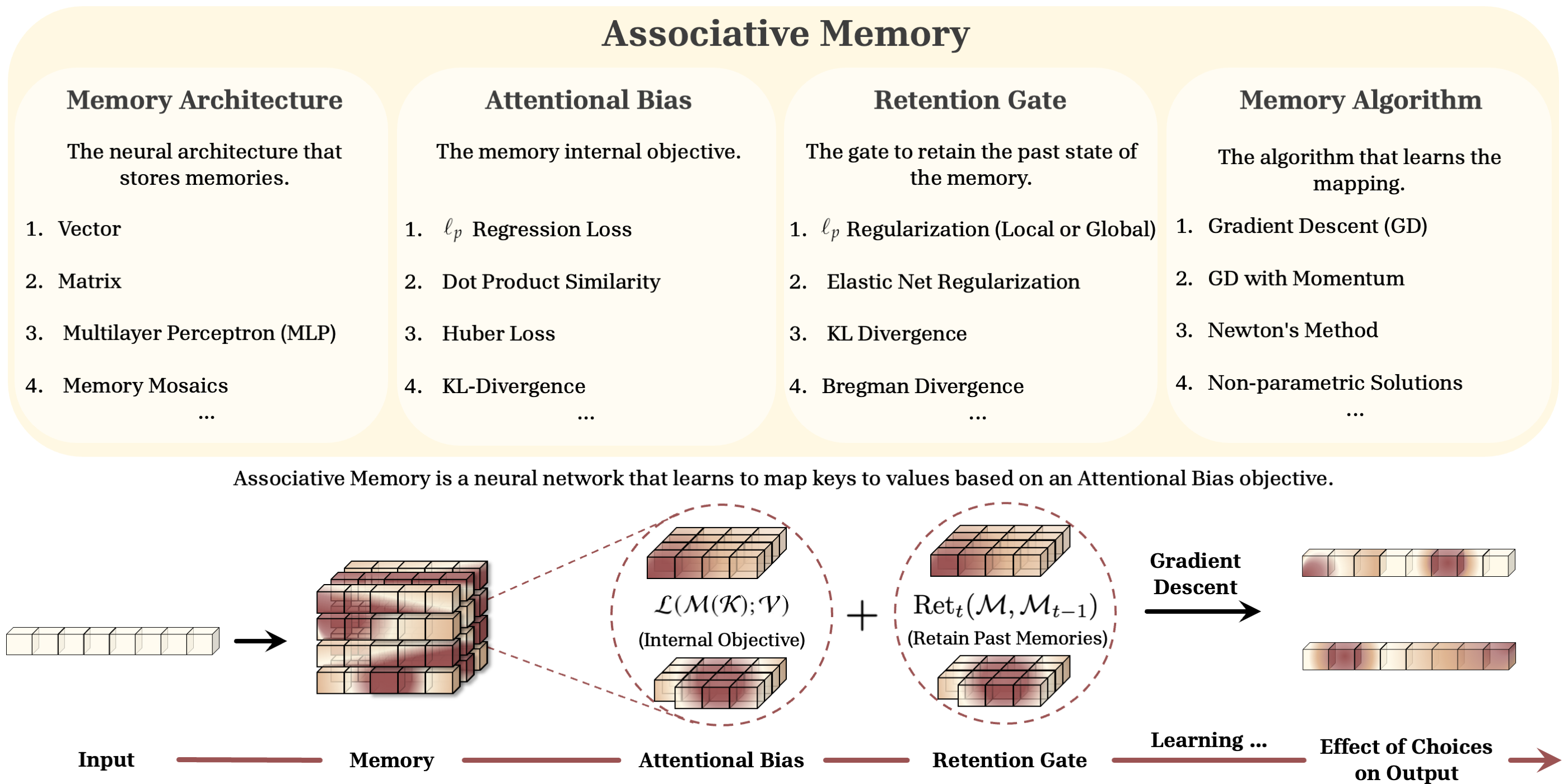}
    \caption{The overview of \framework{} framework. \framework{} is based on four critical choices of (1) memory architecture, (2) attentional bias, (3) retention gate, and (4) memory learning algorithm. In this framework, the memory architecture determines the model capacity to memorize; attentional bias is responsible for modeling the underlying mapping patterns; retention gate determines how to balance learning new concepts and the retention of previously learned concepts; and memory learning algorithm is responsible for memory management.}
    \label{fig:MIRAS-framework}
\end{figure*}

\begin{table}[t!]
    \centering
    \caption{Overview of recent sequence models in \framework{} framework perspective. Surprisingly, all models are using the same type of attentional bias and regularization (forget gate). Note that these architectural choices does not uniquely identify the backbone as there are other design choices (e.g., input-dependency, channel-wise parameters, etc.) as well as the use of other components such as attention, convolutions, etc. Note that for attentional bias and retention gate, we are referring to the original design of \framework, discussed in \autoref{eq:attentional-bias-loss} and \textcolor{c2}{Remark}~\ref{remark:regularization}.}
    \resizebox{\linewidth}{!}{
    \begin{tabular}{l c c c c c c}
    \toprule
    \multirow{2}{*}{Model} & Memory & \multirow{2}{*}{Attentional Bias} & Retention  & {Memory} & \multicolumn{2}{c}{\multirow{2}{*}{Memory Write Operation}} \\
    & Architecture & & Gate$^\dagger$ & Algorithm & & \\
    \midrule
    \midrule
    \multicolumn{7}{c}{Shallow Memory}\\
    \midrule
         RetNet (\citeyear{sun2023retentive})     & Vector         & Dot-Product & L2 &  GD & \multicolumn{2}{l}{$\M_t =   \alpha \M_{t-1} +  {\vvv_t  \vk_t^\top}$}\\
         \midrule
         Transformer (\citeyear{Vaswani+2017}) & Matrix & L2 & - & Nonparametric & \multicolumn{2}{l}{$\M_t = \M_{t-1} \cup \{(\vk_t, \vvv_t)\}$} \\
         LA (\citeyear{schlag2021linear})          & Matrix         & Dot-Product & - &  GD & \multicolumn{2}{l}{$\M_t = \M_{t-1} + \vvv_t \vk_t^{\top}$}\\
         DFW          & Matrix         & Dot-Product & L2 &  GD & \multicolumn{2}{l}{$\M_t = \left(\beta_t \alpha_t^{\top} \right) \odot \M_{t-1} + \vvv_t \vk_t^{\top}$}\\
         Lightening Attention~(\citeyear{li2025minimax})       & Matrix     & Dot-Product & L2 &  GD & \multicolumn{2}{l}{$\M_t = \alpha   \M_{t-1}   +  \vvv_t\vk_t^\top$}\\
         GLA (\citeyear{yang2024gatedattn})        & Matrix     & Dot-Product & L2 &  GD & \multicolumn{2}{l}{$\M_t = \text{Diag}(\alpha_t )  \M_{t-1}   +  \vvv_t\vk_t^\top$}\\
         Mamba (\citeyear{gu2024mamba})     & Matrix         & Dot-Product & L2 &  GD & \multicolumn{2}{l}{$\M_t =   \alpha_t \M_{t-1} + \vvv_t \vk_t^\top $}\\
         HGRN2 (\citeyear{qin2024hgrn})      & Matrix  & L1 & L2 &  GD & \multicolumn{2}{l}{$\M_t =   \text{Diag}( \alpha_t) \M_{t-1}  +  \vvv_t(\mathbf{1} - \alpha_t)^\top $}\\
         DeltaNet (\citeyear{neil2017delta}) & Matrix         & L2 & - &  GD & \multicolumn{2}{l}{$\M_t =   (\mathbf{I} - \beta_t \vk_t \vk_t^\top) \M_{t-1} + \beta_t \vvv_t  \vk_t^\top$}\\
         Longhorn (\citeyear{liu2024longhorn})  & Matrix         & L2 & - &  Implicit GD & \multicolumn{2}{l}{$\M_t =  \left(\mathbf{I}- \frac{\beta_t \vk_t \vk^\top}{\mathbf{1}+ \beta_t \vk_t^\top \vk_t} \right) \M_{t-1} + \left(\frac{\beta_t}{\mathbf{1}+ \vk_t^\top \vk_t \beta_t } \odot \mathbf{x}_t\right) \vk_t$}\\
         TTT-Linear (\citeyear{sun2024learning}) & Matrix         & L2 & - &  GD & \multicolumn{2}{l}{$\M_t = \M_{t-1} - \eta\nabla \mathcal{L}(\M_{t-1}, \mb{x}_t)$}\\
         Gated DeltaNet (\citeyear{yang2024gated}) & Matrix & L2 & L2 & GD & \multicolumn{2}{l}{$\M_t =   \left(\alpha_t (\mathbf{I} - \beta_t \vk_t \vk_t^\top)\right) \M_{t-1} + \beta_t \vvv_t  \vk_t^\top$} \\
         RWKV-7 (\citeyear{peng2025rwkv7}) & Matrix & L2 & L2 & GD & \multicolumn{2}{l}{$\M_t = \text{diag}(\alpha_t)\left( \mathbf{I} - \beta_t \vk_t \vk_t^\top \right) \M_{t-1} + \beta_t \vvv_t  \vk_t^\top$}\\
         DeltaProduct (\citeyear{siems2025deltaproduct}) & Matrix & L2 & L2 & MGD$^*$ & \multicolumn{2}{l}{$\M_t =   \left(\alpha_t \prod_{i = 1}^{n} (\mathbf{I} - \beta_{t, i} \vk_{t, i} \vk_{t, i}^\top)\right) \M_{t-1} + \sum_{j = 1}^{n} \prod_{i = j}^{n} (\mathbf{I} - \beta_{t, i} \vvv_{j, i} \vk_{j, i}^\top)$}\\
         \midrule
         \multicolumn{7}{c}{Deep Memory}\\
         \midrule
         TTT-MLP (\citeyear{sun2024learning})    & 2-layer MLP & L2 & - &  GD & \multicolumn{2}{l}{$\M_t = \M_{t-1} -\eta \nabla \mathcal{L}(\M_{t-1}; \vk_t, \vvv_t)$}\\
         \multirow{1}{*}{Titans-LMM (\citeyear{behrouz2024titans})} & \multirow{1}{*}{$k$-layer MLP} & \multirow{1}{*}{L2} & \multirow{1}{*}{L2} & \multirow{1}{*}{GD + Momentum} & \multicolumn{2}{l}{$\M_t = \alpha_t \M_{t-1} - \mathcal{S}_t, \:\:\: \text{where} \:\: \mathcal{S}_t = \eta_t \mathcal{S}_{t-1} - \theta_t \nabla \mathcal{L}(\M_{t-1}; \vk_t, \vvv_t)$}\\
         \midrule
       \mmodel{}  (\textcolor{c1}{ours}) & $2$-layer MLP & L$_p$ & L$_q$ & GD &  \multicolumn{2}{l}{$A_t = \alpha_t A_{t-1} - \eta_t \nabla \ell_p(W_{i-1};\vk_t, \vvv_t), W_t = \frac{A_t}{\|A_t\|_q^{q-2}}$}\\
        \multirow{2}{*}{\ymodel{}  (\textcolor{c1}{ours})} & \multirow{2}{*}{$2$-layer MLP} & \multirow{2}{*}{Huber} & \multirow{2}{*}{L2} & \multirow{2}{*}{GD} &  \multicolumn{2}{l}{\multirow{2}{*}{$ W_t = \alpha_t W_{t-1} - \begin{cases} 
    \eta_t \: \nabla \ell_2 (W_{t-1}; \vk_t, \vvv_t)   &  \text{if} \quad \| \M(\vk_t) - \vvv_t \| \leq \delta_t, \\
    \eta_t \: \delta_t \nabla  \ell_1 (W_{t-1}; \vk_t, \vvv_t)  & \text{Otherwise}.
    \end{cases} $}} \\
        &&&&&& \\
         &&&&&& \\
        \xmodel{}  (\textcolor{c1}{ours}) & $2$-layer MLP & L2 & KL & GD &  \multicolumn{2}{l}{$W_t = \mbox{Softmax}\left( \alpha_t \log(W_{t-1})  - \eta_t  \nabla \ell_2(W_{t-1};\vk_t, \vvv_t)\right)$}\\
    \toprule
    \multicolumn{7}{l}{$^*$ is using multiple rounds of GD per token. }\\
     \multicolumn{7}{l}{$^\dagger$ For the sake of clarity, we use L2 for all modified L2-like regularizations. However, in fact, only Titans and RWKV-7 are using L2 retention gate (see \autoref{sec:miras})}
    \end{tabular}
    }
    \label{tab:memory-perspective-all}
\end{table}

\section{Preliminaries and Background}\label{sec:background}
In this section, we review the related studies and background concepts that we use through the paper.

\head{Attention}
Attention as the backbone of Transformers is a critical component that acts as their associative memory~\citep{bietti2023birth}. Given input $x \in \R^{N \times d_{\text{in}}}$, causal attention computes output $\mathbf{y} \in \R^{N \times d_{\text{in}}}$ based on \texttt{Softmax} over input dependent key, value, and query matrices:
\begin{align}
    &\mathbf{Q} = x \mathbf{W}_{\mathbf{Q}}, \qquad \mathbf{K} = x \mathbf{W}_{\mathbf{K}}, \qquad \mathbf{V} = x \mathbf{W}_{\mathbf{V}}, \\
    &\mathbf{y}_i = \sum_{j = 1}^{i} \frac{ \exp\left( \mathbf{q}_i^{\top} \mathbf{k}_j/\sqrt{d_{\text{in}}}\right) \mathbf{v}_j }{\sum_{\ell = 1}^{i} \exp\left( \mathbf{q}_i^{\top} \mathbf{k}_{\ell}/\sqrt{d_{\text{in}}}\right)},
\end{align}
where $\mathbf{W}_{\mathbf{Q}}, \mathbf{W}_{\mathbf{K}},$ and $\mathbf{W}_{\mathbf{V}} \in \R^{d_{\text{in}} \times d_{\text{in}}}$ are learnable parameters. While Transformers achieve significant improvements compared to traditional Recurrent Neural Networks (RNNs)—such as LSTM~\citep{LSTM}, their complexity that requires at least $N\times d$ operators to calculate the output has been the main motivation for researchers to think about alternative architectures. We divide and review the research efforts to design alternative architectures into two groups: (1) Linear shallow memory recurrent models, (2) Deep memory modules.

\head{(Linear) Recurrent Models}
For many years, non-linear (gated) recurrent neural networks had been the de facto architectural backbones in deep learning~\citep{greff2016lstm}. Their recurrent nature, however, results in non-parallelizable training, making their large scale training infeasible. To this end, in recent years, linear RNNs as alternatives to both Transformers and non-linear RNNs attract much attention mainly due to their parallelizable and linear-time training while maintaining competitive performance~\citep{yang2024parallelizing, sun2023retentive, rwkv-repo}. Earlier variants of linear RNNs~\citep{yang2024gatedattn, sun2023retentive, de2024griffin}, which mostly are based on Hebbian learning rule~\citep{hebb2005organization}, aim to compress the data into their vector-valued (or matrix-valued) memory~\citep{katharopoulos2020transformers, sun2023retentive, yang2024gatedattn, de2024griffin, liu2024longhorn}. Let $\M_t \in \mathbb{R}^{d \times n}$ be the memory ($n=1$ means vector-valued memory), and $\vk, \vvv \in \mathbb{R}^{d}$ are keys and values (i.e., projection of input $x_t \in \mathbb{R}^{d}$), a simple general formulation for such linear RNNs can be written as:
\begin{align}
    \M_t = A_{t} \ast \M_{t-1} + \vvv_t \vk_t^{\top}, 
\end{align}
where $\ast$ is an arbitrary associative operator and $A_t$ is a data-(in)dependent diagonal matrix or a scalar~\citep{yang2024parallelizing}. Despite the efficiency that comes with the \emph{linear} recurrent nature of these models, the memory can overflow mainly due to the additive (without replacement) nature of Hebbian learning rule, resulting in limited memory capacity and limited expressive power in in-context learning tasks. Moreover, the vector-valued memory of these architectures can limited their ability to learn/memorize large context window, mainly due to the limited expressive power of memory to learn the underlying patterns of data~\citep{behrouz2024titans, sun2024learning}.
 
To address the above mentioned limitations, recurrent models that use a matrix-valued memory with Delta learning rule has gained popularity in recent years~\citep{neil2017delta,schlag2021linear, yang2024parallelizing}. Despite significant advantages, even these delta-rule-based recurrent models face theoretical limitations~\citep{irie2023practical} with moderate performance in practice~\citep{yang2024parallelizing}. Recently, several studies aim to improve the performance of such models by adding scalar or channel-wise forget gate mechanisms~\citep{yang2024gated, peng2025rwkv7}, , using negative eigenvalues~\citep{grazzi2024unlocking}, and multiple learning steps~\citep{siems2025deltaproduct}. They, however, still suffer from performance drop in long context, mainly due to the less expressive memory architectures~\citep{behrouz2024titans}.

\head{Deep Memory Module: Titans and Test Time Training}
To overcome the limited memory and to extend the \emph{effective} context length of deep sequence models, more recent studies focus on a new generation of architectures with deep memory module~\citep{behrouz2024titans, sun2024learning}. These architectures are built on the meta-learning perspective, where the memory is an MLP architecture that is updated using gradient descent (with momentum)~\citep{behrouz2024titans, sun2024learning}. \citet{sun2024learning} further provide a unifying perspective that how linear and softmax attention are respectively parametric and non-parameteric solutions of (kernel) regression loss but consider other modern linear RNNs outside of this class of models. Recently, in a concurrent work to ours, \citet{wang2025test} show that with additional simplification of modern RNNs (e.g., RetNet~\citep{sun2023retentive}, Mamba~\citep{dao2024transformers}) they approximately place in the same class of models that internally optimize regression loss. It, however, still remains unanswered that ``What is the underlying design framework behind these sequence models that can \emph{accurately} unify existing architectures?'' Moreover, the role of forget gates and its alternative choices in modern sequence models is surprisingly less explored.



\section{Associative Memory, Attentional Bias, and Retention} \label{sec:AM}
Associative memory, which is an inseparable component of learning in humans~\citep{terry2017learning}, has been the inspiration for many artificial neural architectures in the literature~\citep{hopfield1982neural, neil2017delta, behrouz2024titans}. These studies, however, define instances of the concept of associative memory, limiting the architecture to a specific class of similarity metrics between entities (i.e., keys and values). That is, broadly speaking, associative memory is an operator that maps a set of keys $K$ to a set of values $V$, and so to learn the underlying mapping patterns in data, it requires an objective that targets a type of memory and measures the quality of learned mappings:

\begin{definition}[Associative Memory and Attentional Bias] \label{dfn:associative-memory}
Given a set of keys $\mathcal{K} \subseteq \mathbb{R}^{d_k}$ and values $\mathcal{V} \subseteq \mathbb{R}^{d_v}$, associative memory is an operator $\M: \mathcal{K} \rightarrow \mathcal{V}$. Learning the mapping of associative memory is based on an objective $\mathcal{L}$, called \emph{Attentional Bias}, that determines the type of memory and its tendency to prioritize some events:
\begin{align}\label{eq:attentional-bias-loss}
    \M^* = \arg\min_{\M}\quad \mathcal{L}(\M(\mathcal{K}); \mathcal{V}).
\end{align}
\end{definition}
A few remarks are in order:
\begin{remark}\label{remark:regularization}
 When we parameterize the memory with parameter $W$, we use $\M(W,\vk)$. In this parameteric setting,  the optimization problem in~\eqref{eq:attentional-bias-loss} should be performed over the parameter $W$. Furthermore, in the parametric setup, we might use an additional regularization $\mathcal{R}(W)$ to control the retaining of the past data.
\end{remark} 

\begin{remark}
Learning the mapping between keys and values (\autoref{eq:attentional-bias-loss}) is a meta-learning problem, in which the attentional bias is optimized in the inner-loop and all other parameters of the neural network (e.g., linear projections, convolutions, etc.) are optimized in the outer-loop. Therefore, the model learns how to store the data into its parameters at test time~\citep{sun2024learning, behrouz2024titans}.
\end{remark}

\subsection{Learning to Memorize and to Retain Through the Lens of Optimization}
\autoref{dfn:associative-memory} translates the design of a neural architecture based on the concept of associative memory to learning the underlying mapping between keys and values, by minimizing an objective $\mathcal{L}$. To optimize~\autoref{eq:attentional-bias-loss}, one simple approach is to utilize the idea of gradient descent. Specifically, given a new pair of keys and values, we update the memory as:
\begin{equation} \label{eq:OGDupdate} 
    W_t = W_{t-1} - \eta_t \nabla \ell (W_{t-1};\vk_t, \vvv_t),  
\end{equation}
where, for simplicity, we use the definition $\ell (W_{t-1};\vk_t, \vvv_t) := \mathcal{L}(\M(W; \vk_t), \vvv_t$). 
\citet{behrouz2024titans} re-interpreter this formulation as a momentary surprise metric, where the model memorizes tokens that violates the expectation of the objective (i.e., being surprising to the memory). Although the choice of objective is an important step to fully interpret \autoref{eq:OGDupdate} (which we discuss in detail in \autoref{sec:beyond-variants}), there are different viewpoints to interpret this update rule in its general format, which later can help us to go beyond existing architectures:

\subsection{Viewpoint 1: Online Regression and Follow-The-Regularized-Leader}
Equation~\eqref{eq:OGDupdate} can be viewed as one step of online gradient descent over the sequence of the loss functions 
\begin{align}
    \ell(W;\vk_1, \vvv_1), \ell(W;\vk_2, \vvv_2), \ldots, \ell(W;\vk_t, \vvv_t), \ldots.
\end{align}
It is well known that the online gradient descent can be viewed as a special case of Follow-The-Regularized-Leader (FTRL) algorithm with a special choice of loss functions \citep[Chapter~2]{shalev2012online} and \citep{hazan2016introduction}. 
Specifically, assuming $W_0=0$, the update rule in \eqref{eq:OGDupdate} is equivalent to 
\begin{equation} \label{eq:OGDasFTRL}
W_{t} = \arg\min_{W}\quad \sum_{i=1}^t \langle W-W_{i-1} , \nabla \ell(W_{i-1};\vk_i, \vvv_i) \rangle  + \frac{1}{2\eta} \|W\|_2^2,
\end{equation}
where the term $\langle W-W_{i-1} , \nabla \ell(W_{i-1};\vk_i, \vvv_i) \rangle$ is the local linear approximation of the original loss at time $i$ and the second term is a regularization term.  While the first part~$ \sum_{i=1}^t \langle W-W_{i-1} , \nabla \ell(W_{i-1};\vk_i, \vvv_i) \rangle$ measures how well can the memory learn all the past tokens, the second term~
$\frac{1}{2\eta} \|W\|_2^2$ penalizes the memory update with respect to the size of memory. 

Equation~\eqref{eq:OGDasFTRL} uses linear approximation of the loss function and quadratic regularization. We can, however, in principle use other approximations of the loss function as well as other regularization functions, as used in the past in online optimization~\citep{shalev2012online,hazan2016introduction} or in general optimization~\citep{mairal2015incremental,razaviyayn2013unified}. Such changes are the idea behind the development of other optimization algorithms such  mirror descent. More specifically, we can generalize the update rule in \eqref{eq:OGDasFTRL} to the form:

\begin{equation}
\label{eq:Viewpoint1-OnlineRegression}
\tag{FTRL Viewpoint}
W_{t} = \arg\min_{W\in \cW}\quad \undermath{\text{Attentional Bias}}{\sum_{i=1}^t \widehat{\ell}_i(W;\vk_i, \vvv_i)}  + \undermath{\text{Memory Stability}}{\frac{1}{\eta_t} \mathcal{R}_t(W)}.
\end{equation}

In this update rule, the  term $\sum_{i=1}^t \widehat{\ell}_i(W;\vk_i, \vvv_i)$ aims at memorizing the tokens at test time, while the term $\mathcal{R}_t(W)$ regularizes the learning dynamics and take the size of the memory into account when updating it by a new incoming data. 
Choosing different loss functions $\widehat{\ell}_i(W;x_i)$ and the regularization term~$\frac{1}{\eta_t} \mathcal{R}_t(W)$ can lead to different algorithms such as (online) gradient descent or  mirror descent. In this generalization,  $\eta_t$ to can be data-dependent. Moreover, we will allow imposing constraint $\cW$ on the choice $W$.

\subsection{Viewpoint 2: Learning the Latest Token While Retaining Previous Information}
Another way to interpret the update rule \eqref{eq:OGDupdate} is to view it as learning from the latest key-value pair $(\vk_i, \vvv_i)$ (via using its gradient or surprise metric), while staying close to the previous state $W_{t-1}$ to retain the previously memorized tokens. Formally, \eqref{eq:OGDupdate} is equivalent to
\[
W_{t} = \arg\min_{W}\quad  \langle W-W_{t-1} , \nabla \ell(W_{t-1};\vk_t, \vvv_t) \rangle  + \frac{1}{2\eta_t} \|W - W_{t-1}\|_2^2
\]
The first term locally approximates $\ell(W;\vk_t, \vvv_t)$ around the previous state $W_{t-1}$, while the last term regularizes  deviations from $W_{t-1}$. This form can generalize to 

\begin{align}
\label{eq:Viewpoint2-Loss-Premetric}
\tag{Learning-Retaining Viewpoint}
  W_t = \arg\min_{W\in \cW} \:\: \undermath{\text{Attentional Bias}}{\widetilde{\ell}_t(W;\vk_t, \vvv_t)} + \undermath{\text{Retention}}{ \mbox{Ret}_t\left(W,W_{t-1}\right)},
\end{align}
where the term $\widetilde{\ell}_t(W;\vk_t, \vvv_t)$ is an approximation of $\ell(W;\vk_t, \vvv_t)$ and minimizing it corresponds to \textit{Learning} from the new concepts~$(\vk_t, \vvv_t)$.  The second term $\mbox{Ret}_t\left(W,W_{t-1}\right)$ regularizes the changes in $W$ to make the learning dynamics stable and to \textit{retain} previously learned knowledge. This Retention function may have local and global components:
\[
\mbox{Ret}_t\left(W,W_{t-1}\right) = \undermath{\text{Local Retention}}{\frac{1}{\eta_t} \mbox{D}_t\left(W,W_{t-1}\right)} + \undermath{\text{Global Retention}}{\frac{1}{\alpha_t} \mbox{G}_t\left(W\right)}.
\]
Here, the term $\mbox{D}_t\left(W,W_{t-1}\right)$, which is a \textit{premetric} that controls the deviations from $W_{t-1}$, aims at \textit{retaining} previously learned knowledge. The coefficient $\eta_t$ can be viewed as a meta in-context learning rate, where larger values of $\eta_t$ leads to learning more from new concepts, while allowing higher forgetting of previously learned concepts.  The second term is a global retention that controls the change of the memory with respect to its size. The special instances of the above viewpoint (e.g., \emph{without} global retention, with implicit \emph{closed-form} solution, and/or with limited memory structure) have been the motivation behind some of the recent studies such as~\citet{liu2024longhorn}.

\subsection{Further Discussions on the Two Viewpoints}
The \eqref{eq:Viewpoint1-OnlineRegression} and \eqref{eq:Viewpoint2-Loss-Premetric} are connected through the lens of online optimization. For example, as discussed above, by choosing linear approximation of the loss and quadratic regularization/retention, they can both cover online gradient descent update in~\eqref{eq:OGDupdate} as a special case.  One straightforward way to make the connection  explicit is by defining the premetric $\mbox{D}_t(W;W^\prime)$ based on the previous loss functions and the regularization, as described in Proposition~\ref{prop:Viewpoint1Implies2} below:
\begin{proposition}
\label{prop:Viewpoint1Implies2}
 Let $\eta_t = \eta$ and define~$h_t(W) :=  \sum_{i=1}^{t-1} \widehat{\ell}_i (W;\vk_i,\vvv_i) + \frac{1}{\eta}R(W)$. Assume $\cW = \mathbb{R}^d$ and the function $h_t(W)$ is strictly convex in $W$ and let $\mathcal{D}_h(\cdot,\cdot)$ be the Bregman divergence defined by function $h(\cdot)$, i.e., $\mathcal{D}_h(W,W^\prime) = h(W) - h(W^\prime) - \langle \nabla h(W^{\prime}), W-W^{\prime}\rangle$. Set $\mbox{Ret}_t(W,W^\prime) = \mathcal{D}_h(W,W^\prime)$ and $\widetilde{\ell}_t(W;x_t) = \widehat{\ell}_t(W;x_t)$ in \eqref{eq:Viewpoint2-Loss-Premetric}. Then, the update rule in~\eqref{eq:Viewpoint2-Loss-Premetric} is equivalent to the update rule in~\eqref{eq:Viewpoint1-OnlineRegression}. 
\end{proposition} 
We provide the proof in \autoref{app:proof}. The above proposition shows that \eqref{eq:Viewpoint2-Loss-Premetric} can also explain the approaches obtained by~\eqref{eq:Viewpoint1-OnlineRegression}, under some mild assumptions. Hence, \eqref{eq:Viewpoint2-Loss-Premetric} may be seen as a more general version. This is why we focus on this viewpoint in most of our derivations in the next sections.

\begin{remark}
Given the above viewpoint, we can see that even by using additional global regularization there is no memory erasing or forgetting process (a common term in modern architectures~\citep{behrouz2024titans, yang2024gated}) but the model might decide to not retain the past state of the memory. Interestingly, this observation also matches the human memory process, where brain does not erase memories but they might become inaccessible due to retrieval failures~\citep{robertson2002memory}. Therefore, instead of calling it a forget gate, later on, we use ``Retention Gate'' to refer to this term.  
\end{remark}

\begin{remark}
As we discuss in \autoref{sec:miras} and summarize in \autoref{tab:memory-perspective-all}, most existing modern sequence models are optimizing associative memory objective (attentional bias in \autoref{eq:attentional-bias-loss}) using gradient descent. Therefore, to provide further intuition about the connection of existing sequence models as well as their online learning interpretations, we discuss the above two viewpoints that are limited to gradient descent-based update rules. Our initial definition of attentional bias and associative memory in \autoref{eq:attentional-bias-loss}, however, is broader and can be optimized by any optimization algorithm (e.g., even Newton's method, or non-parametric solutions).  
\end{remark}

\section{\framework: Learning to Memorize with Robust and Expressive Memory}\label{sec:miras}

Building upon our definition of associative memory, attentional bias, and previous viewpoints, we present \framework{} framework that not only \emph{accurately} unifies existing backbone architectures but it also provides insights on how to design the next generation of sequence models. As discussed earlier in \autoref{sec:AM}, learning an associative memory can be interpreted as a meta-learning task, in which the associative memory learns how to compress and store data into its parameters at test time. The architecture of the memory in such tasks is particularly important as in longer contexts, the expressivity of the memory structure can limit its ability to learn the underlying patterns. Therefore, the first choice to design a sequence model is the structure of the memory. 
Given the structure of the memory, parameterized by a set of parameters $W$, as discussed earlier, we aim to minimize a loss function $\ell(W; \cdot, \cdot)$ with a retention regularizer $\mbox{Ret}(\cdot)$ via a learning algorithm (e.g., gradient descent). Accordingly, \framework{} requires four design~choices:
\begin{enumerate}[leftmargin=15pt]
    \item \textbf{Memory Structure:} This choice specifies the architecture of the memory. For example, this architecture can be a vector, a linear function, a Multilayer Perceptron (MLP) layer, or even more complex structures. We may restrict the choice of $W$ to be within a certain region, e.g., $W$ to lie within an $L_2$ ball to avoid infinite values or unstable training.
    \item \textbf{Attentional Bias:} A key choice is the attentional bias objective~$\mathcal{L}(\cdot)$ in \autoref{eq:attentional-bias-loss}. We can even consider different approximations of the loss function, (e.g.,  $\widehat{\ell} (\cdot,\cdot)$ in \eqref{eq:Viewpoint1-OnlineRegression} or $\widetilde{\ell}(\cdot,\cdot)$ in \eqref{eq:Viewpoint2-Loss-Premetric}). The choice of attentional bias determines how memory memorizes the context, maps the inputs, and prioritizes the events.
    \item \textbf{Memory Stability and Retention:} Another key choice is the retention regularizer $\mathcal{R}(\cdot)$ (e.g., $\mathcal{R}_t(\cdot)$ in \eqref{eq:Viewpoint1-OnlineRegression} and $\mbox{Ret}_t(\cdot)$ in \eqref{eq:Viewpoint2-Loss-Premetric}). In parametric setups, this choice balances learning with retention of past state. An effective retention gate is key to the good performance in~long~context tasks. 
    \item \textbf{Memory Algorithm:} Finally, this choice specifies the learning algorithm that we use to optimize the memory objective. One may use gradient descent, gradient descent with momentum, or any other algorithm (including finding non-parametric solutions).  
\end{enumerate}
The above choices are major design choices for designing backbone sequence models in neural architectures. There are, however, minor decisions that can distinguish models; i.e., data-dependent or independent parameters, scalar or channel-wise learning rate/retaining gate, etc. Next, we discuss the overview of how existing architectures fit into \framework{} framework. 
 
\head{RNNs with Hebbian Rule} The first generation of modern recurrent architectures (e.g., Linear attention~\citep{katharopoulos2020transformers}, RetNet~\citep{sun2023retentive}, Mamba~\citep{gu2024mamba}, and GLA~\citep{yang2024gatedattn}) are based on Hebbian-like (e.g., gated Hebbian) learning rule~\citep{hebb2005organization}. We let attentional bias be the dot product similarity. That is, given a memory $\M \in \R^{d \times n}$ and $\vk, \vvv \in \R^{d}$, we define $\tilde{\ell}_t := -2 \inner{\M_t \vk_t}{\vvv_t}$ and \emph{local retention} as $\text{Ret}_t(\M, \M_{t-1}) = \| \M_t - \alpha \M_{t-1} \|_F^2$. Using \autoref{eq:Viewpoint2-Loss-Premetric} and gradient descent as the optimizer (i.e., memory learning algorithm), the memory update rule is:
\begin{align}
    \M_t = \alpha \M_{t-1} + \vvv_t \vk_t^{\top}.
\end{align}
When (1) $\alpha = 1$, memory update is equivalent to Linear Attention (LA)~\citep{katharopoulos2020transformers}; (2) $\alpha \in \R$ is a learnable parameter, resulting architecture is either lightening attention ($n > 1$)~\citep{li2025minimax} or RetNet ($n = 1$)~\citep{sun2023retentive}; and (3) $\alpha_t \in \R$ are \emph{data-dependent} learnable parameters, resulting sequence model is Mamba2~\citep{dao2024transformers}.

\head{RNNs with Delta Rule} To improve the memory management and to enhance the memory capacity of the above group, several studies suggest using delta rule~\citep{neil2017delta, schlag2021linear} as the learning algorithm in recurrent neural networks (e.g., DeltaNet~\citep{schlag2021linear}, Longhorn~\citep{liu2024longhorn}, and RWKV7~\citep{peng2025rwkv7}). In this part, we recall that where $\M \in \R^{d \times n}$, delta rule is equivalent to optimizing MSE objective $\|\M_t \vk_t - \vvv_t\|^2_2$ with $\text{Ret}_t(\M, \M_{t-1}) = \| \M_t - \alpha \M_{t-1} \|_F^2$ as local retention, and stochastic gradient descent as optimizer: ($\eta_t$ is defined in \autoref{eq:Viewpoint2-Loss-Premetric})
\begin{align}
    \M_t = \alpha \left( \mathbf{I} - \eta_t \vk_t \vk_t^{\top}\right) \M_{t-1} + \vvv_t \vk_t^{\top}.
\end{align}
When (1) $\alpha = 1$, memory update is equivalent to DeltaNet~\citep{schlag2021linear}; and (2) $\alpha_t \in \R^m$ are \emph{data-dependent} learnable parameters, resulting sequence model is either Gated DeltaNet~\citep{yang2024gated} ($m = 1$), or RWKV7~\citep{peng2025rwkv7} ($m = d$).  
Therefore, RNNs with delta rule are special instances of \framework.

\head{Beyond Delta Rule}
As discussed earlier, while delta rule with its value replacement strategy is more powerful than Hebbian-like learning rules, it suffers from theoretical limitations~\citep{irie2023practical} and achieves moderate performance in practice~\citep{yang2024parallelizing}. Therefore, several studies have focused on update rules beyond delta rule. Recently, Titans~\citep{behrouz2024titans} suggests using non-linear MSE objective of $\|\M_t(\vk_t) - \vvv_t\|^2_2$ with both local and global retention of $\mbox{D}_t = \|W_t - W_{t-1} \|_F^2$ and $\mbox{G}_t = \|W_t \|^2_2$ and optimize it with gradient descent with \emph{momentum} \footnote{The retention gate (forget gate) in Titans is different from Mamba2 and Gated DeltaNet that we discussed above. The main difference comes from the case of full memory erase. While Mamba2 gating removes the entire memory and treats the next token as the first ever seen data, Titans use a ``\emph{cold start}'' strategy and use the previous state of the memory to measure the surprise of the incoming token before fully erasing the memory.}. Therefore, Titans-LMM is a special instance of \framework, where we use the abovementioned attentional bias and retention regularizations, and gradient descent with momentum as the optimizer. 

Another example of such models is Mesa-layer, in which the model uses $\sum_{i = 1}^{t} \|\M_t(\vk_i) - \vvv_i\|^2_2$ as the attentional bias objective with $\|\M_t\|_2^{2}$ as the retention regularization. Since these models uses Newton's method to optimize such an objective, they provide a more expressive update rule than delta rule. We further discuss a set of new learning algorithms beyond delta rule in \autoref{sec:beyond-variants}.

\head{Attention}
As discussed by \citet{sun2024learning}, softmax attention is a non-parameteric solution of $\ell_2$-MSE loss function (i.e., $\|W \vk - \vvv\|_2^2$) with Nadaraya-Watson estimator. Therefore, softmax attention is an instance of \framework, when we find the non-parameteric solution to the MSE loss with Nadaraya-Watson estimator, without retention.

\section{Beyond Existing Attentional Biases and Retention Gates}\label{sec:beyond-variants}
As discussed in the previous section, existing work focus only on linear/quadratic choices for the attentional bias or retention gate. In particular, the loss function~$L(\M(\vk_t),\vvv_t)$ is defined as $L(\M(\vk_t),\vvv_t) = c_t\|\M(\vk_t)-\vvv_t\|^2$ for some (learnable) constant $c_t$ in prior work. Also the regularization term $R_t(W)$ or the parametric $\mbox{D}_t$ is considered as a quadratic/linear function. In addition, almost all prior work consider $W$ to be the entire $\mathbb{R}^d$ space. However, in general there could be various choices for all the three aforementioned design choices. To illustrate the potential and flexibility of our designed framework, here, we review some of the potential design choices for attentional bias and retention gate in \framework. For the sake of clarity, we discuss all these attentional bias and memory retention gates based on using gradient descent as the optimizer, and so based on the provided two view points. However, these attentional bias objectives and retention regularizers can be directly used in \autoref{eq:attentional-bias-loss} and optimized by using any other optimization algorithms, resulting in different update rules.

\subsection{Alternative Attentional Biases}

\head{Variant 1: $\ell_p$-Attentional Bias}
As discussed in the main body, attentional bias defines the ``similarity metric'' and measures how well memory can recall the value, given its corresponding key. Although $\ell_2$ regression loss often is a natural choice, it is sensitive to noise in the data. A natural extension is to use $\ell_p$-norm class of objectives. That is, let $\M$ be the memory, $\vk$ be the keys, and $\vvv$ be the values, we define $\ell_p$-attentional bias as:
\begin{align}
    \mathcal{L}(\M(W, \vk_t); \vvv_t) = \| \M(\vk_t) - \vvv_t \|^p_p,
\end{align}
where $p \in \mathbb{R}^{\geq 1}$ and $\|.\|_p$ is the $p$-norm. Although depending on the distribution of the data, we might want to use different values of $p$ (see \autoref{sec:exp}), different values of $p$ can result in memory architectures with interesting properties. For the sake of simplicity, let memory be a matrix, i.e., $W \in \R^{m \times d}$ and $\M(W, \vk_t) = W \vk_t$, the closed form can be derived as:
\begin{align}\label{eq:lp-attentional-bias}
    W_t = W_t - \eta_t \nabla \ell (W_{t-1}; \vk_t, \vvv_t) = W_t - p \: \eta_t \:\: \left(\mbox{Sign}(W \vk_t - \vvv_t) \odot|W \vk_t - \vvv_t|^{p-1}\right)\:\vk_t^{\top}.
\end{align}
Let $p = 1$, the recurrence is simplified as:
\begin{align}
    W_t =  W_t - \eta_t \:\: \mbox{Sign}(W_t \vk_t - \vvv_t) \:\vk_t^{\top},
\end{align}
which means that the memory has only two values of $-1$ and $1$. We call this variation \emph{value-less} associative memory, in which we store entities (keys) but map them into two extreme class of -1 and +1.

\begin{remark}
One of the critical challenges to use the above update rule is in the backpropagation process, in which $\mbox{Sign}(\cdot)$ and $|\cdot|$ are non-differentiable and so might cause unstable training. To overcome this issue, we use $\mbox{Sign}(x) \approx \mbox{tanh}\left( \alpha x \right),$ and $|x| = \sqrt{x^2 + \epsilon},$ as the smooth approximators of these functions.  
\end{remark}

One simple interpretation for such behavior (i.e., value-less memory) is similar to the coping mechanism in humans~\citep{loftus1993reality}, in which the memory does not store the values for extreme events. This interpretation of protective memory in extreme events motivates our next variant.

\head{Variant 2: Huber Loss: Memory with Coping Mechanism} 
While $\ell_2$-norm  objective is a common choice for many statistical and machine learning tasks, it is known to be sensitive to outliers and extreme samples.This sensitivity extends to the use of $\ell_2$ loss for attentional bias.
To address this and drawing motivation from robust regression literature, we suggest utilizing the Huber loss-type~\citep{huber1992robust, hastie2009elements} as the attentional bias, thereby reducing the negative impact of the outlier data on the memory learning process. 

We can apply Huber-type loss in three different ways: The first approach is to define the summation of the Huber loss across different coordinates as the total loss, i.e., 
\[
\ell(W;\vk_t,\vvv_t) = \sum_{j}\mathcal{H}(\M(W,\vk_t)_{j}-\vvv_{t,j}),
\]
where $\M(W,\vk_t)_{j}$ and $\vvv_{t,j}$ denote the $j$-th coordinate of $\M(W,\vk_t)$ and $\vvv_t$ respectively. The function $\mathcal{H}(\cdot):\mathbb{R}\mapsto \mathbb{R}$ is the Huber loss defined as
\begin{equation}
\mathcal{H}(a) = \left\{
\begin{array}{lc}
    \frac{1}{2}a^2 & \textrm{if}\; |a|\leq \delta  \\
    \delta \left(|a|-\frac{1}{2}\delta\right) & \textrm{if}\; |a|> \delta.
\end{array}
\right.
\end{equation}
Utilizing this attentional bias can lead to various memory update rules. For example, for the matrix form memory $\M(W,\vk_t) = W\vk_t,$ the update rule is given by
\begin{align}
    W_t = W_{t-1} - \eta_t \left[
    \left((W\vk_t - \vvv_t)\vk_t^T\right)\odot \left(\mathbf{I}(|W\vk_t - \vvv_t| \leq \delta_t) \mathbf{1}^\top\right)
    + 
     \left(\delta_t \mbox{Sign}(W\vk_t - \vvv_t) \vk^\top\right)\odot \left(\mathbf{I}(|W\vk_t - \vvv_t| > \delta_t) \mathbf{1}^\top\right)   
    \right]
\end{align}
In this formulation, the parameter $\delta_t$ decides the type of the memory used for each block of memory ($\ell_2$-norm objective or value-less) based on the context, making the memory more robust to outliers.

The second approach is to define the Huber-type loss based on the $\ell_2$ loss over all coordinates, i.e., 
\[
\ell(W;\vk_t,\vvv_t)  = \mathcal{H} (\|\M(W,\vk_t) - \vvv_t\|_2).
\]
For simplicity of derivations, assume matrix memory $M(W,\vk_t) = W\vk_t$. Then using gradient descent for updating memory leads the memory update rule
\begin{align}
\label{eq:temp42}
    W_t = W_{t-1} - \eta_t \: 
    \begin{cases} 
     \left(\M(W_{t-1},\vk_t)-\vvv_t\right)\vk_t^T   &  \text{if} \quad \| \M(W_{t-1},\vk_t) - \vvv_t \|_2 \leq \delta_t, \\
   \delta_t \frac{\left(\M(W_{t-1},\vk_t)-\vvv_t\right)}{\|\M(W_{t-1},\vk_t)-\vvv_t\|_2}\vk_t^T   & \text{Otherwise}.
    \end{cases} 
\end{align}
Again, in the form~\eqref{eq:temp42}, the parameter $\delta_t$ decides the type of the memory used  ($\ell_2$-norm objective or normalized version) based on the context, making the memory more robust to outliers. 

Finally, in the third approach, we present a smooth mixture method, in which the memory decides if for an incoming data it is better to use $\ell_2$ or $\ell_1$ attentional bias: 
\begin{align}\label{eq:huber-loss}
    W_t = W_{t-1} - \begin{cases} 
    \eta_t \: \nabla \ell_2 (W_{t-1}; \vk_t, \vvv_t)   &  \text{if} \quad \| \M(\vk_t) - \vvv_t \| \leq \delta_t, \\
    \eta_t \: \delta_t \nabla  \ell_1 (W_{t-1}; \vk_t, \vvv_t)  & \text{Otherwise}.
    \end{cases} 
\end{align}
The role of parameter $\delta_t$ is the same as above.

\head{Variant 3: Memory Robust to Value Shifts}
Following the robustness requirement discussed in the previous section, we aim to design a memory mechanism that exhibits resilience against small shifts in the value parameter. A natural approach in this context is to employ a robust optimization formulation. Specifically, we define the loss function as the worst-case $\ell_2$ distance between the predicted memory output and the perturbed true value:
\begin{align}
\label{eq:tempRobustRegression}
    \mathcal{L}(\M(W, \vk_t); \vvv_t) = \max_{\|\boldsymbol{\delta} \vvv_t\|_2 \leq \Delta }\frac{1}{2}\| \M(W,\vk_t) - (\vvv_t+\boldsymbol{\delta} \vvv_t) \|_2^2.
\end{align}
This formulation seeks the memory parameters $W$ that perform well even under the  adverse local perturbation of the true value $\mathbf{v}_t$ within an $\ell_2$ ball of radius $\Delta$.
To solve the maximization problem in \eqref{eq:tempRobustRegression}, we find the optimal perturbation $\boldsymbol{\delta}\mathbf{v}_t^*$. By solving this problem with respect to $\boldsymbol{\delta}\mathbf{v}_t$, we arrive at:
\[
\boldsymbol{\delta}\vvv_t^* = \Delta \frac{-\M(W,\vk_t) + \vvv_t}{\|\M(W,\vk_t) - \vvv_t\|_2}
\]
Substituting this optimal perturbation back into the loss function \eqref{eq:tempRobustRegression}, we obtain the robust loss:
\[
\mathcal{L}(\M(W, \vk_t); \vvv_t) = \frac{1}{2}\|\M(W,\vk_t) - \vvv_t\|_2^2 + \Delta \|\M(W,\vk_t) - \vvv_t\|_2 + \frac{1}{2}\Delta^2.
\]
This robust loss function is a combination of the standard $\ell_2$ loss and a term proportional to the $\ell_2$ norm of the error, scaled by the robustness parameter $\Delta$. The value of $\Delta$ thus controls the trade-off between fitting the nominal data and ensuring robustness against value perturbations.

For simplicity of the derivations, let us consider a constant value for $\Delta$, an Euclidean retention gate  $\mbox{Ret}_t\left(W,W_{t-1}\right) = \|W-W_{t-1}\|^2$, and an attentional bias term $\widetilde{\ell}(W;\mathbf{k}_t, \mathbf{v}_t) = \langle W-W_{t-1}, \nabla \ell(W_{t-1};\mathbf{k}_t, \mathbf{v}_t)\rangle$. Furthermore, to simplify the memory operation, we assume a linear matrix memory model $\M(W,\mathbf{k}_t) = W\mathbf{k}_t$. Under these assumptions, we can derive the memory update mechanism using gradient descent on the robust loss:

\[
W_{t} = W_{t-1} - \eta \left( \big(\M(W_{t-1},\vk_t)-\vvv_t\big) \vk_t^\top + \Delta \frac{\M(W_{t-1},\vk_t)-\vvv_t}{\|\M(W_{t-1},\vk_t)-\vvv_t\|_2}\vk_t^\top
\right)
\]
In this update rule, the parameter $\Delta$, which governs the influence of the robustness term, can also be treated as a learnable parameter, allowing the model to adapt its robustness based on the observed data.

\subsection{Alternative Retention Gates}\label{app:variants}

\head{Variant 1: Memorization Over A Scaled Probability Simplex Via $f$-Divergence} 
A common technique in learning to prevent numerical instabilities and exploding values is to restrict the search space to a bounded domain. Following this principle, to avoid numerical instabilities,  we can constrained the variable $W_t$ to lie within a (scaled) probability simplex. In other words, we can restrict the state to lie in the constraint set 
\begin{align*}
    \cW = \{W\, \mid \, \|W\|_1 = c \; \textrm {and} \; W_{jl} \geq 0, \;\forall j,l\}.
\end{align*}
In this set, each matrix~$W$ can be viewed as a measure. Thus, in \eqref{eq:Viewpoint2-Loss-Premetric}, we can utilize divergences over measures to define our premetric. For example,  we can use $f$-divergence measure ~\citep[Def 4.9]{polyanskiy2025information}, \citep{csiszar1967information} to define $\mbox{D}_t(\cdot,\cdot)$. More specifically, let $f(\cdot)$ be a smooth strictly convex function from $\mathbb{R}^+$ to $\mathbb{R}$ with $f(1)=0$. Then, we can define the $f-$ divergence between $W$ and $W'$ as
\[
\mbox{D}_t (W,W^\prime) = \sum_{jl} W^{\prime}_{jl} \, f\left(\frac{W_{jl}}{W^{\prime}_{jl}}\right).
\]
It is known that $f$-divergence is zero if and only if $W = W^\prime$; see~\cite[Theorem 2.3]{polyanskiy2025information}. Using the above premetric as the retention gate and setting $\widetilde{\ell}(W;\vk_t, \vvv_t) = \langle W-W_{t-1}, \nabla \ell(W_{t-1};\vk_t, \vvv_t)\rangle$ in \eqref{eq:Viewpoint2-Loss-Premetric}, we get the update rule
\begin{equation}
\label{eq:generalf-divergence}
W_t = W_{t-1} \odot g\left( -\zeta_t - \eta_t \nabla \ell(W_{t-1};\vk_t, \vvv_t)\right).
\end{equation}
Here $g(\cdot)$ is the inverse of the mapping $f^{\prime}$, i.e., $g(f' (\tau)) =\tau,\;\forall \tau$; the operator $\odot$ denotes the Hadamard (elementwise) product, and $\zeta_t$ should be chosen such that $\|W_t\|_1 = c$. Notice that since the function $f(\cdot)$ is strictly convex and smooth, its derivative is strictly increasing and hence $g(\cdot)$ is well defined. Conversely, for any strictly monotone  function $g(\cdot)$, we can find its inverse function $g^{-1}$ (which is strictly increasing)  and define $f(\tau) =  \textrm{const}+\int_{\tau' = 0}^{\infty} g^{-1}(\tau') d\tau'$. The term $\textrm{const}$ should be chosen such that $f(1) = 0$. Then the update rule in \eqref{eq:generalf-divergence} can be interpreted by the $f$-divergence regularization, as explained above. Therefore, one can directly choose a continuous monotonically increasing function $g(\cdot)$ and use \eqref{eq:generalf-divergence} for memory update. 

\vspace{0.2cm}

\textbf{Specializing to KL divergence.} Let us further make the above update rule explicit by using special  function $f$. If we choose $f(\tau) = \tau \ln(\tau)$, then the $f$-divergence becomes the widely used KL divergence measure~$D_t(W,W_{t-1}) = \sum_{jl} W_{jl}\log\left(\frac{W_{jl}}{(W_t)_{jl}}\right)$. 
In addition, we can also utilize the Shannon entropy as the global retention by regularizing deviations from uniform distribution, i.e., $G_t(W) =  \sum_{jl}W_{jl}\log(W_{jl})$.  
Combining these choices of the local and global retention gates, we obtain the overall retention gate
\[
\mbox{Ret}_t(W,W_{t-1}) = \frac{1}{\eta_t} \sum_{jl} W_{jl}\log\left(\frac{W_{jl}}{(W_t)_{jl}}\right) + \frac{1}{\alpha_t} \sum_{jl}W_{jl}\log(W_{jl})
\]
Choosing the attentional bias~$\widetilde{\ell}(W;\vk_t, \vvv_t) = \langle W-W_{t-1}, \nabla \ell(W_{t-1};\vk_t, \vvv_t)\rangle$ and the above retention gate will lead to the  update rule
\begin{align}
W_{t} = \arg\min_{W} \;&\langle W-W_{t-1}, \nabla \ell(W_{t-1};\vk_t, \vvv_t)\rangle +  \frac{1}{\eta_t} \sum_{jl} W_{jl}\log\left(\frac{W_{jl}}{(W_t)_{jl}}\right) + \frac{1}{\alpha_t} \sum_{jl}W_{jl}\log(W_{jl}) \\
\textrm{s.t.}\quad & \sum_{jl}W_{jl} = c,\; W_{jl} \geq 0, \;\forall jl
\end{align}
Attaching the Lagrange multiplier to the first constraint, the KKT conditions implies
\[
\left(\nabla \ell(W_{t-1};\vk_t, \vvv_t)\right)_{jl} +\left(\frac{1}{\eta_t} + \frac{1}{\alpha_t}\right)\left(1+ \log W_{jl}\right) - \frac{1}{\eta_t} \log\left((W_{t-1})_{jl}\right) + \mu_t = 0,\quad \forall j,l
\]
where $\mu_t$ should be chosen such that $ \sum_{jl}W_{jl} = c$. Rearranging the terms and defining~$\lambda_t = \frac{1/\alpha_t}{1/\alpha_t + 1/\eta_t}$, $\eta'_t = \frac{1}{1/\alpha_t + 1/\eta_t},$  we get the update rule
\begin{equation}\label{eq:xmodel}
     W_t \leftarrow c \; \mbox{Softmax}\left( (1-\lambda_t) \log(W_{t-1})  - \eta'_t \nabla \ell(W_{t-1};\vk_t, \vvv_t)\right)
\end{equation}
where $\lambda_t \in (0,1)$ and $\eta' \in \mathbb{R}^+$ are the parameters that can be learned during training. The Softmax operator ensures that the output lies in the set $\cW$.  

Notice that while all above calculations are done for a matrix $W$, similar update rule holds for other forms of parameters such as when $W$ is a neural network (or when the parameter~$W$ is normalized per slice).

\head{Variant 2: Elastic Net Regularization: Hard and Soft Forgetting} 
Elastic net is a powerful and popular tool in regression analysis to balance the feature selection capabilities of LASSO \citep{tibshirani1996regression} and bias reduction properties of Ridge regression \citep{hilt1977ridge,hoerl1970ridge}. It has been widely used  in different applications due to its ability to handle high-dimensional data and mitigate the effects of multicollinearity. Given this success, a natural question is what happens if we use this regularization scheme in our context. 

Let us start based on \eqref{eq:Viewpoint2-Loss-Premetric} to design our memorization scheme. In \eqref{eq:Viewpoint2-Loss-Premetric}, we discussed that the loss function $\widetilde{\ell}_t(W;\vk_t, \vvv_t)$ is an approximation of the original function~$\ell(\cdot)$, measuring our goodness-of-fit. Regularizing this loss with elastic net regularizer, we obtain the approximation
\[
\widetilde{\ell}_t(W;\vk_t, \vvv_t)= \langle W-W_{t-1},\nabla \ell(W_{t-1};\vk_t, \vvv_t) \rangle.
\]
with a global retention of $\mbox{G}_t(W) = \frac{1}{2\beta} \|W\|_2^2 + \frac{1}{\alpha} \|W\|_1$. 
To fully specify the update rule of \eqref{eq:Viewpoint2-Loss-Premetric}, we also need to specify the premetric functions~$\mbox{D}_t(\cdot,\cdot)$. For the sake of keeping the update rule simple (and parallelizable), we can choose
\[
\mbox{D}_t(W,W_{t-1}) = \frac{1}{2}\|W-W_{t-1}\|_2^2.
\]
These choices of the attentional bias and retention gate leads to the following update rule:

\begin{equation}\label{eq:Elastic}
    W_t = \mathcal{S}_{\gamma} \left( \lambda W_{t-1} - \zeta \nabla \ell(W_{t-1};\vk_t, \vvv_t)\right),
\end{equation}
where $\gamma =  \frac{\eta \beta}{\alpha(\eta + \beta)}$, $\lambda = \frac{\beta}{\beta+\eta}$, $\zeta = \eta \lambda$, and $\mathcal{S}_{\gamma}$ is the soft thresholding operator, applied element-wise. For each element,  this operator is defined as
\[
 \mathcal{S}_{\gamma}(z) = \mbox{sign}(z) \max\left\{0, |z| - \gamma\right\}.
\]
In other words, for large values of $z$, $ \mathcal{S}_{\gamma}(z)$ makes $z$ closer to zero by $\gamma$ amount. If it is already in the $\gamma$-vicinity of zero, then it makes it zero (hard forget). 

Equation~\eqref{eq:Elastic} can be viewed as a combination of soft forgetting (obtained by multiplying $W$ by $\lambda \in (0,1)$, and a hard forgetting (if it is smaller than $\gamma$). The hyperparameters $\gamma$, $\lambda$, and $\zeta$ can be learned.
Notice that since the shrinkage operator is not  differentiable, we can  approximate it with its smooth approximation. For example, we can use
$\mathcal{S}_{\gamma}(z) \approx \frac{|z| * \arctan(z / \gamma)}{\pi / 2}$.

\head{Variant 3: Elastic Net Regularization: Forgetting via Soft-thresholding} The elastic net regularizer can also be used in the \eqref{eq:Viewpoint1-OnlineRegression}. In particular, in \eqref{eq:Viewpoint1-OnlineRegression}, we can set
\[
\frac{1}{\eta_t}R_t(W) = \frac{1}{\eta} \|W\|^2 + \frac{1}{\alpha} \|W\|_1
\]
and use $\widehat{\ell}(W;x_i) = \langle W-W_{i-1}, \nabla \ell(W_{i-1};x_i)\rangle$. Assuming initialization at $W_0 = 0$, these choices of attentional bias and retention gate leads to the update rules:
\begin{align}
    A_t &= A_{t-1} - \eta \nabla \ell(W_{t-1};\vk_t, \vvv_t) \nonumber\\
    W_t &= \mathcal{S}_{\eta/\alpha} \left(A_t\right)
\end{align}
Here $\mathcal{S}_{\eta/\alpha} (\cdot)$ is the soft-thresholding operator with parameter $\eta/\alpha$, which can be smoothly as explained in Variant~1.1.

\head{Variant 4: General $L_q$ Memory Stability} Existing work is based on the retention gate choices $\mbox{D}_t(W,W_{t-1}) = \|W-W_{t-1}\|^2_F$ or $R(W) = \|W\|_2^2$. However, one can choose other choices of retention gate. For example, in \eqref{eq:Viewpoint1-OnlineRegression}, we can choose $L_q$ norm as the regularizer $R(W)$. More specifically, for $1<q\leq 2$, we can set 
\[
\frac{1}{\eta_t}R(W) = \frac{1}{2\eta (q-1)} \|W\|_q^2.
\]
Using this retention gate and choosing $\widehat{\ell}_i(W;\vk_t, \vvv_t) = \langle W-W_{i-1} , \nabla \ell(W_{i-1};\vk_t, \vvv_t)  \rangle$ in \eqref{eq:Viewpoint1-OnlineRegression},  leads to the update rule
$
W_t = - \eta \frac{A_t}{\|A_t\|_p^{p-2}},
$
where $p = \frac{q}{q-1}$ and $A_t = \sum_{i=1}^t\nabla \ell(W_{i-1};\vk_t, \vvv_t)$; see \cite[Section 2.6]{shalev2012online}. Here, $\odot$ denotes the Hadamard (element-wise) product and $|\cdot|$ is the element-wise absolute value operator.  Assuming $W_0 =0$, this update rule can be recursively written as:
\[
A_t = A_{t-1} - \eta \nabla \ell(W_{i-1};\vk_t, \vvv_t),
\quad
\textrm{and}
\quad
W_t = \frac{A_t}{\|A_t\|_p^{p-2}}.
\]

\head{Variant 5:  Bregman Divergence as Retention Gate.} Another natural choice is to use Bregman divergence as retention gate, leading to a mirror descent-type algorithms. In particular, given a smooth strictly convex function $f(\cdot):\mathbb{R} \mapsto \mathbb{R}$, we can define the function $F(W) = \sum_{jl} f(W_{jl})$. Based on this choice of function $F$, we define the Bregman divergence
\[
D_t(W,W') = F(W) - F(W') - \langle W', W-W'\rangle
\]
 as our parametric function. Utilizing this retention gate and choosing $\widetilde{\ell}_t(W;\vk_t, \vvv_t) = \langle W-W_{t-1} , \nabla \ell(W_{t-1};\vk_t, \vvv_t)  \rangle$
 in \eqref{eq:Viewpoint2-Loss-Premetric}, we obtain the update rule 
 \[
 W_t = g\left(-\eta \nabla \ell(W_{t-1};\vk_t, \vvv_t) + F'(W_{t-1})\right).
 \]
Here, $F'$ is the mapping obtained by applying $f'(\cdot)$ (the derivative of $f$) element-wise to all entries of its  input matrix argument. The function $g$ is the inverse of the mapping $F'(\cdot)$, i.e., $g(F'(W)) = W$.

If we choose $f(\tau) = \frac{\tau^2}{2}$, then $F'(W)$ becomes the identity mapping and so is $g$. Therefore, the above update becomes simple gradient descent with no nonlinearity involved in the update rule. However, other choices of $f(\cdot)$ introduces additional nonlinearity in $g(\cdot)$, which can enhance the expressivity of our memory. For example, we can choose the function $f(\cdot)$ so that its derivative becomes the inverse sigmoid function, i.e., $f'(\tau) = \ln\left(\frac{\tau}{1-\tau}\right)$ with $f':(0,1)\mapsto\mathbb{R}$. Since $f'(\cdot)$ is strictly increasing, then the function $f(\cdot)$ (and hence $F(\cdot)$) is strictly convex. Therefore, the Bregman divergence is well defined. Moreover, the inverse of the function $f'(\cdot)$ becomes the sigmoid function, i.e., $g(\tau) = \sigma(\tau) = \frac{\exp(\tau)}{1+\exp(\tau)}$ with $g:\mathbb{R}\mapsto (0,1)$. Then, the update of the memory becomes
\[
 W_t = \sigma\left( \ln \left(\frac{W_t}{1-W_t}\right)-\eta \nabla \ell(W_{t-1};\vk_t, \vvv_t) \right),
\]
where $\sigma$ is the sigmoid function operated element-wise on the entries of $W$, and the division operator $\frac{W_t}{1-W_t}$ is also performed element-wise. This update rule guarantees that the elements of $W_t$ remains within the interval $(0,1)$.

\subsection{\framework's Variants: \mmodel, \ymodel, and \xmodel}\label{sec:Miras-variants}
In the previous section we discussed different potential choices for attentional bias and retention gate to show the generality and the potential of \framework. In this section, building upon our framework, we present three novel sequence models, each of which designed based on a different motivation, and discuss how they can leverage fast parallel training.

\head{\mmodel}
Given $p, q \in\R^{\geq 1}$, we design $(p, q)$-\mmodel{} as the variant of \framework{} as follows: (1) For the choice of memory architecture, we use an MLP with 2 layers with expansion factor of 4 and GELU activation function~\citep{hendrycks2016gaussian}. We also use residual connections and layer norm, resulting in $\M(x) = x + \texttt{LN}(W_1 \sigma(W_2 x))$. (2) We choose $\ell_p$-attentional bias (introduced in \autoref{eq:lp-attentional-bias}) for \mmodel. (3) For the choice of retention gate, we use the hybrid of $\ell_q$ retention gate $\frac{1}{2(q-1)} \|W\|_q^2$ (see \autoref{app:variants} for details) and the standard $\ell_2$ regularization $\frac{1}{\beta} \|W\|_2^2$. (4) Finally, we use gradient descent as the memory learning algorithm. The above choices, result in the following recurrent formula for the memory module:
\begin{align}\label{eq:q-norm}
    A_t = \alpha_t A_{t-1} - \eta_t \nabla \ell_p(W_{i-1};\vk_t, \vvv_t),
\quad
\textrm{and}
\quad
W_t = \frac{A_t}{\|A_t\|_q^{q-2}}.
\end{align}
Notably the gradient can be calculated using:
\begin{align}
   \nabla \ell (W_{t-1}; \vk_t, \vvv_t) =  p \: \eta_t \:\: \left(\mbox{Sign}(W \vk_t - \vvv_t) \odot|W \vk_t - \vvv_t|^{p-1}\right)\:\vk_t^{\top}.
\end{align}
We use $(p, q)=(3, 4)$.

\head{\ymodel}
Building upon our discussion on the importance of robust memory that protects itself from extreme events (tokens), we design \ymodel{} based on Huber objective. That is, in \framework, for the choice of memory structure, we follow \mmodel{} and use an MLP with the same architecture as above; for the choice of attentional bias, we use Huber loss (defined in \autoref{eq:huber-loss}); for the choice retention gate, for the sake of simplicity, we use a combination of local and global retention as $\mbox{Ret}_t (W, W_{t-1}) = \frac{1}{2 \theta_t} \|W - W_{t-1}\|_{\text{F}}^2 + \frac{1}{\beta_t} \|W\|_2^2$, which is equivalent to the ``forget gate'' mechanism introduced by \citet{behrouz2024titans}; and finally, we simply use gradient descent as the memory learning algorithm.  Given the above choices, we can write the resulted memory learning process as follows:
\begin{align}
    W_t = \alpha_t W_{t-1} - \begin{cases} 
    \eta_t \: \nabla \ell_2 (W_{t-1}; \vk_t, \vvv_t)   &  \text{if} \quad \| \M(\vk_t) - \vvv_t \| \leq \delta_t, \\
    \eta_t \: \delta_t \nabla  \ell_1 (W_{t-1}; \vk_t, \vvv_t)  & \text{Otherwise}.
    \end{cases} 
\end{align}
Note that for improving the expressive power, in all architectures, we decouple the learning rate $\eta$ and the retention gate rate $\alpha$, resulting in an independent parameter $\beta_t \in [0, 1]^d$.

\head{\xmodel}
Finally, in \xmodel, we use the idea of elastic net regularization (i.e., hard and soft retention). To this end, in \framework: (1) For the choice of memory architecture, similar to above variants, we use an MLP (the same architecture as the previous variants). (2) For the choice of attentional bias, we use simple $\ell_2$ regression loss. (3) For the choice of retention gate we use KL divergence as in \autoref{eq:xmodel}. (4) Finally, we optimize the memory using gradient descent, resulting in the following update rule: 

\begin{align}
  W_t = \mbox{Softmax}\left( \alpha_t \log(W_{t-1})  - \eta_t  \nabla \ell_2(W_{t-1};\vk_t, \vvv_t)\right)
\end{align}

\begin{figure*}
    \centering
    \includegraphics[width=0.85\linewidth]{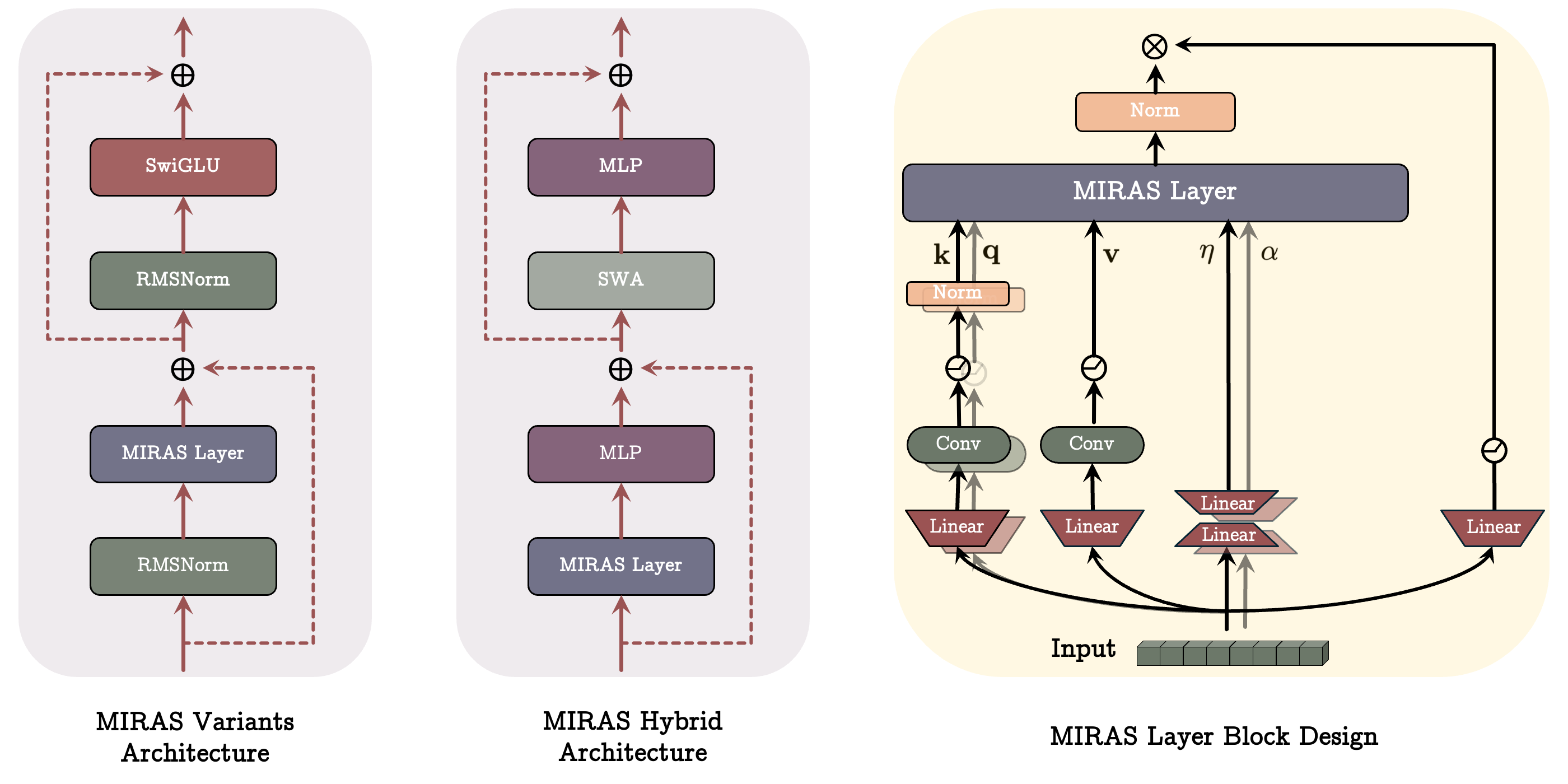}
    \caption{Visualization of the \framework's variant architecture, their hybrid counterpart with SWA, and block design of \framework{} layer.}
    \label{fig:arch-vis}
\end{figure*}

\subsection{Architecture Backbone and Fast Training}

\head{Architectural Backbone} For the architectural backbone, we fully follow recent studies~\citep{behrouz2024titans, yang2024gated}: We replace attention modules with our variants of \framework{} in Llama’s macro architecture with MLPs with \texttt{SwiGLU}(.) activation, rotary positional encodings (RoPE)~\citep{su2024roformer}, and RMSNorm~\citep{zhang2019root}. For \framework{} layer block, we follow the recent modern linear recurrent models~\citep{yang2024gated, behrouz2024titans}, and incorporate a 1D depthwise-separable convolution layer (with kernel size of 4) after each of the query, key, and value projections. For the sake of training stability, we also use $\ell_2$ normalization to $\vq$ and $\vk$. The output of \framework{} layer block is normalized and gated with a linear layer~\citep{mehta2023long}. 

\head{Channel-wise Parameters} For learnable parameters of $\eta_t, \delta_t$ and the retention gate of $\alpha_t$ we use channel-wise parametrization, i.e., $\eta_t, \delta_t, \alpha_t \in \R^{d}$. While gaining more expressive power, this parametrization results in significant parameter increase. To mitigate this issue, following \citet{peng2025rwkv7}, we use low-rank projections to project the input into $\R^k$ and then to $\R^d$, where $k$ is a hyperparameter (usually 32 or 64). The backbone architecture is illustrated in \autoref{fig:arch-vis}. 

\head{Hybrid Models} We also evaluate the hybrid version of \framework's variants. For hybrid models, we follow the Samba~\citep{ren2024samba} architecture, in which we sequentially combine our \framework{} layer with Sliding Window Attention (SWA). The illustration of hybrid model \autoref{fig:arch-vis}. 

\head{Parallelizable Training}
While the design of \framework's variant are theoretically well-motivated, their recurrence is non-linear, potentially make their straightforward training slow for large scales. In this section, we build upon the work of \citet{behrouz2024titans, sun2024learning} to make the training parallelizable. The main idea is to divide the sequence into chunks with size $b$ (usually is 16 or 64) and calculate the gradient for all tokens in the current chunk with respect to the last state of the memory in the previous chunk. That is, we use $\nabla \ell(\M_{t'}; \vk_t, \vvv_t)$ instead of $\nabla \ell(\M_{t-1}; \vk_t, \vvv_t)$, where $t'$ is the last state in the previous chunk. 

Given the above trick, we can calculate all gradients at once and make the recurrence inside each chunk linear. However, to fully take advantage of accelerators, we need to reformulate the process as matrix multiplication. For \mmodel{}, for the sake of clarity, assume $q = 2$. We follow the same algorithm as \citet{behrouz2024titans} and expand the recurrence as follows:
\begin{align}\nonumber
    \M_{t} &= \alpha_t \M_{t-1} - \eta_t \nabla \ell(\M_{t-1}; \vk_t, \vvv_t) \\
    &= \beta_{t} \M_0 - \sum_{i = 1}^{t} \eta_i \frac{\beta_{t}}{\beta_{i}} \nabla \ell(\M_{t'}; \vk_i, \vvv_i),
\end{align}
where $t' = t - \texttt{mod}(t, b)$, and $\beta_{i} = \prod_{j=1}^{i}\alpha_j$. For the sake of clarity, we focus on the first chunk, i.e., $t = b$ and so $t' = 0$, and explain the process for the case that $\M_{t} = W_t$ is linear. The process for 2-layer MLPs and other chunks is similar. Using $\ell_p$ loss function, we have:
\begin{align}\nonumber
    &\nabla \ell(W_{0};\vk_t, \vvv_t) =  p \left(\mbox{Sign}(W_0 \vk_t - \vvv_t) \odot|W_0 \vk_t - \vvv_t|^{p-1}\right)\:\vk_t^{\top} \\ \label{eq:weight-decay-matmul}
    \Rightarrow &\sum_{i = 1}^{b} \eta_i \frac{\beta_{b}}{\beta_{i}} \nabla \ell(W_{0}; ;\vk_i, \vvv_i)  = p  \mathbf{E}_b \odot \mathbf{B}_b \odot \mbox{Sign}(W \vk_t - \vvv_t) \odot (|W_0 \: \mathbf{K} - \mathbf{V}|^{p-1}) \: \mathbf{K}^{\top}, 
\end{align}
where $\mathbf{E}_b = \begin{bmatrix}
    \eta_1 & \eta_2 & \dots & \eta_b
\end{bmatrix}$ and $\mathbf{B}_b$ is defined analogously on $\frac{\beta_{b}}{\beta_{i}}$s. For the sake of stablity in training, we use $\texttt{Sign}(x) \approx \texttt{tanh}\left( \alpha x \right)$ and $|x| = \sqrt{x^2 + \epsilon}$, where $\epsilon > 0$ is a small number (i.e., $\epsilon = 1e-6$). As discussed in \autoref{eq:q-norm}, the case that $q \neq 2$ appears as a normalization term on the memory. Similar to Titans~\citep{behrouz2024titans} and TTT~\citep{sun2024learning}, we do not apply this non-linearity inside each chunk and instead use it at the end of each chunk. 

For \ymodel, the process is very similar to the above. We calculate the gradient of both $\ell_1$ and $\ell_2$ loss and use a masking based on $\|\M(\vk_t) - \vvv_t\| \leq \delta_t$. 

For \xmodel, the update rule has two non-linear part, i.e., softmax and $\log$, making the model hardly parallelizable. To this end, as discussed above, we use its linear version inside each chunk and its non-linear version across chunks. However, using both log and softmax at the end of each chunk removes the effect of log. To this end, we consider a lag tokens after each chunk (i.e., tokens with index $i = k b + 1$, where $b$ is the chunk size and $k \in \mathbb{Z}^{+}$). That is, let $\M_0$ be the last state of the memory in previous chunk, we have:
\begin{align}
    \M_1 = \mbox{Softmax}\left( \alpha_1 \log(\M_{0})  - \eta_1  \nabla \ell_2(\M_{0};\vk_1, \vvv_1)\right),
\end{align}
and then we use $\M_1$ for the next chunk. Again, for the sake of clarity, assume that memory is linear, i.e.,  $\M_1 = W_1$:
\begin{align}
    &\nabla \ell(W_{1};\vk_t, \vvv_t) = (W_1 \vk_t - \vvv_t)\:\vk_t^{\top} \\ \label{eq:weight-decay-matmul2}
    \Rightarrow &\sum_{i = 1}^{b} \eta_i \frac{\beta_{b}}{\beta_{i}} \nabla \ell(W_{1}; ;\vk_i, \vvv_i)  =  \mathbf{E}_b \odot \mathbf{B}_b \odot (W_1 \: \mathbf{K} - \mathbf{V}) \: \mathbf{K}^{\top}, 
\end{align}
where matrices are defined the same as for \autoref{eq:weight-decay-matmul}.

\section{Experiments} \label{sec:exp}
In our experimental evaluations, we aim to answer three main questions: (1) Does different attentional biases results in different architectures in practice? (2) How does different types of retention gates (i.e., retention gate) affect the performance of the model in long context? (3) How do \xmodel, \mmodel, and \ymodel{} perform in downstream tasks compare to baselines?

\head{Setup}
We train our models with training context window of size 4096 using either FineWeb-Edu dataset~\citep{penedo2024the} (for LM and common-sense reasoning tasks) or C4 dataset~\citep{raffel2020exploring} (for scaling patterns). We use model sizes of 120M, 340M, 760M, and 1.3B parameters. We train small models (120M and 340M) on 15B tokens sampled from the dataset, the medium size model (760M) on 30B tokens, and the large model on 100B tokens. Baseline results are reported by \citet{behrouz2024titans}.

\subsection{Language Modeling and Common-sense Reasoning}
We follow recent studies~\citep{yang2024gated, yang2024parallelizing, behrouz2024titans} and first focus on the perplexity in language modeling and also commonsense reasoning tasks. The results for \xmodel, \ymodel, \mmodel{} and also baselines with size of 340M, 760, and 1.3B are reported in \autoref{tab:lm_results}. All of our variants outperforms all the baselines including Transformer++, modern linear recurrent models and hybrid methods. The superior performance compared to hybrid models is particularly important as all of our variants are pure recurrent (attention-free). Among the three variants of \framework, while \mmodel{} achieves slightly weaker performance than \xmodel, and \ymodel, the other two variants are close and depending on the task and model size, the best model can vary.

\begin{table*}[t!]
\centering
\caption{
Performance of \framework's variants and baselines on language modeling and common-sense reasoning tasks. Hybrid models are marked with~$^*$. The best results of \colorbox{myblue}{simple} and \colorbox{mygreen}{hybrid} models are highlighted. In largest scale, we compare our simple models with even hybrid models and \colorbox{myblue}{highlight} the best results.
}\label{tab:lm_results}
\centering
\resizebox{0.9\linewidth}{!}{
\centering
\begin{tabular}{l|c c|c c c c c c c c }
\toprule
\textbf{Model}  & \textbf{Wiki.}  &  \textbf{LMB.} &  \textbf{LMB.} & \textbf{PIQA} &    \textbf{Hella.} & \textbf{Wino.} & \textbf{ARC-e} &  \textbf{ARC-c} &  \textbf{SIQA}  & \textbf{BoolQ} \\
 & ppl $\downarrow$  &  ppl $\downarrow$  &  acc $\uparrow$  & acc $\uparrow$ &   acc\_n $\uparrow$  & acc $\uparrow$  & acc $\uparrow$ & acc\_n $\uparrow$ &  acc $\uparrow$  & acc $\uparrow$  \\
\midrule
\midrule
\multicolumn{11}{c}{340M params / 15B tokens} \\
\midrule
 Transformer++ & 31.52 & 41.08 &  30.76 & 62.98  &  34.76 & 50.53  & 45.21  & 24.05 & 36.81 & 58.24 \\
 RetNet & 32.50 & 49.73 & 28.24 & 62.61 & 34.15 &  50.91 & 44.27 & 23.62 & 36.79 & 59.72 \\
 GLA & 28.51 & 43.02 & 28.73 & 64.05 & 35.96 & 50.00 & 54.19 & 24.29 & 37.13 & 58.39 \\
 Mamba & 30.83 & 40.21 & 29.94 & 63.79 & 35.88 & 49.82 & 49.24 & 24.56 &  35.41  & 60.07 \\
 DeltaNet & 28.65 & 47.30 & 28.43 & 63.52 & 35.95 & 49.63 & 52.68 & 25.37 &  \cellcolor{myblue} {37.96}  &  58.79  \\
 TTT & 27.44 & 34.19 & 30.06 & 63.97  & 35.71 & 50.08 & 53.01 & 26.11 & 37.32 & 59.83 \\
 Gated DeltaNet & 27.01 & 30.94 &  34.11 & 63.08 & 38.12  &  \cellcolor{myblue} 51.60  &  55.28  &  26.77  & 34.89 & 59.54  \\
\midrule
\mmodel{} (ours) & \cellcolor{myblue}26.19 &  29.31 &  \cellcolor{myblue} 35.70  & 63.99  & 39.23  & 52.04  & \cellcolor{myblue} 55.96  &  27.15 & 37.29 &  \cellcolor{myblue} 60.22 \\
\ymodel{} (ours) & 26.61 &  \cellcolor{myblue} 29.11 &  34.09  & 64.93  & \cellcolor{myblue} 39.86  & 51.12  & 54.75  & \cellcolor{myblue} 28.64 & 33.82 & 60.29 \\
\xmodel{} (ours) & 27.16 &  30.44 &  33.68  & \cellcolor{myblue} 65.21  & 39.17  & 51.23  & 53.40  & 27.99 & 34.1& 59.29  \\
\midrule
\multicolumn{11}{c}{760M params / 30B tokens} \\
\midrule
 Transformer++ & 25.21 & 27.64 & 35.78 & 66.92 & 42.19 &	51.95 & 60.38	& 32.46 &  39.51  & 60.37  \\
 RetNet & 26.08 & 24.45 & 34.51 & 67.19 & 41.63 &	52.09 & 63.17	& 32.78 &  38.36  & 57.92  \\
 Mamba2 & 22.94 & 28.37 & 33.54 & 67.90 & 42.71 &	49.77 & {63.48}	& 31.09 &  40.06  & 58.15   \\
 DeltaNet & 24.37 & 24.60 & 37.06 & 66.93 & 41.98 &	50.65 & 64.87	& 31.39 &  39.88  & 59.02   \\
 TTT & 24.17 & 23.51 & 34.74 & 67.25 & 43.92 & 50.99 & 64.53 & 33.81 & {40.16} & 59.58  \\
 Gated DeltaNet & {21.18} & {22.09} & {35.54} & {68.01} & {44.95} & {50.73} & {66.87}	& {33.09} &  {39.21}  & 59.14  \\
 Samba$^{*}$ & 20.63 & 22.71 & 39.72 & 69.19 & 47.35 &	52.01 & 66.92	& 33.20 &  38.98  & 61.24  \\
  Gated DeltaNet-H2$^*$ & {19.88} & 20.83 & {39.18} & 68.95 & {48.22} &	{52.57} & 67.01	& {35.49} &  {39.39}  & 61.11  \\
  \midrule
\mmodel{} (ours) & 21.18 & 21.94 & 38.02 & \cellcolor{myblue} 69.55  & 49.16 & 53.01 & 67.47 & 36.09 & 40.53 & 63.18  \\
\ymodel{} (ours) &  
20.99 & 21.57 & 37.85 & 69.14  & \cellcolor{myblue} 50.02 & \cellcolor{myblue} 53.93 & \cellcolor{myblue} 67.78 & \cellcolor{myblue} 36.27 & \cellcolor{myblue} 41.01 & \cellcolor{myblue} 63.34 \\
\xmodel{} (ours) &  22.28 & 22.31 & 38.19 & 67.82  & 49.30 & 53.28 & 63.57 & 36.15 & 40.94 & 62.96 \\
  \midrule
\mmodel-H (ours) &  18.72 & 20.13  & \cellcolor{mygreen} 40.59   & \cellcolor{mygreen} 70.84 & 50.13 & \cellcolor{mygreen} 54.17  & 67.64  & \cellcolor{mygreen}36.79 & 40.87 & \cellcolor{mygreen}62.43 \\
\ymodel-H (ours) & 18.59 & \cellcolor{mygreen} 19.80  & 40.22   &  69.51 & \cellcolor{mygreen} 50.48 & 53.69  & \cellcolor{mygreen}68.04  & 36.55 & 40.28 & 61.94 \\
\xmodel-H (ours) & \cellcolor{mygreen} 18.24 & 20.55  & 39.91   &  69.06 & 49.84 & 52.88  & 66.90  & 36.12 & \cellcolor{mygreen} 40.99 & 61.75 \\
\midrule
\multicolumn{11}{c}{1.3B params / 100B tokens} \\
\midrule
 Transformer++ & 18.53 & 18.32 & 42.60 & 70.02 & 50.23 &	53.51 & 68.83	& 35.10 &  40.66  & 57.09  \\
 RetNet & 19.08 & 17.27 & 40.52 & 70.07 & 49.16 &	54.14 & 67.34	& 33.78 &  {40.78}  & {60.39} \\
 Mamba2 & {16.56} & {12.56} & {45.66} & {71.87} & {55.67} &	{55.24} & {72.47}	& {37.88} &  40.20  & 60.13   \\
 DeltaNet & 17.71 & 16.88 & 42.46 & 70.72 & 50.93 &	53.35 & 68.47	& 35.66 &  40.22  & 55.29   \\
 Gated DeltaNet & {16.42} & {12.17} & {46.65} & {72.25} & {55.76} &{57.45} & {71.21}	& {38.39} &  {40.63}  & 60.24  \\
 Samba$^*$ & 16.13 & 13.29 & 44.94 & 70.94 & 53.42 &	55.56 & 68.81	& 36.17 &  39.96  & \cellcolor{myblue} {62.11}   \\
  Gated DeltaNet-H2$^*$ & {15.91} &{12.55} & \cellcolor{myblue} {48.76} & {72.19} & \cellcolor{myblue} {56.88} &	{57.77} & {71.33}	 &{39.07} &  {41.91}  & 61.55     \\
  \midrule
\mmodel{} (ours) & 15.52 & \cellcolor{myblue} 11.47  & 47.88   &  \cellcolor{myblue} 73.16 & 56.14 & \cellcolor{myblue} 59.09  & \cellcolor{myblue} 72.53  & \cellcolor{myblue} 40.32 & \cellcolor{myblue} 41.91 & 61.18  \\
\ymodel{} (ours) &  \cellcolor{myblue} 15.18 & 11.89 & 47.23   &  72.81  &  56.46  & 59.02  & 72.14 & 40.05 &  40.73 &  61.86\\
\xmodel{} (ours) &  15.90 & 12.04  &  48.67   &  73.10 &  55.99 &  57.36 & 71.55  & 37.92 & 40.19 & 61.34 \\
\bottomrule
\end{tabular}
}
\end{table*}


\subsection{Scaling Pattern}
To evaluate the scaling pattern of models and for comparing them with baseline, in this section, we plot their performance with varying the model size and the context window. 

\begin{figure*}
    \centering
    \includegraphics[width=0.32\linewidth]{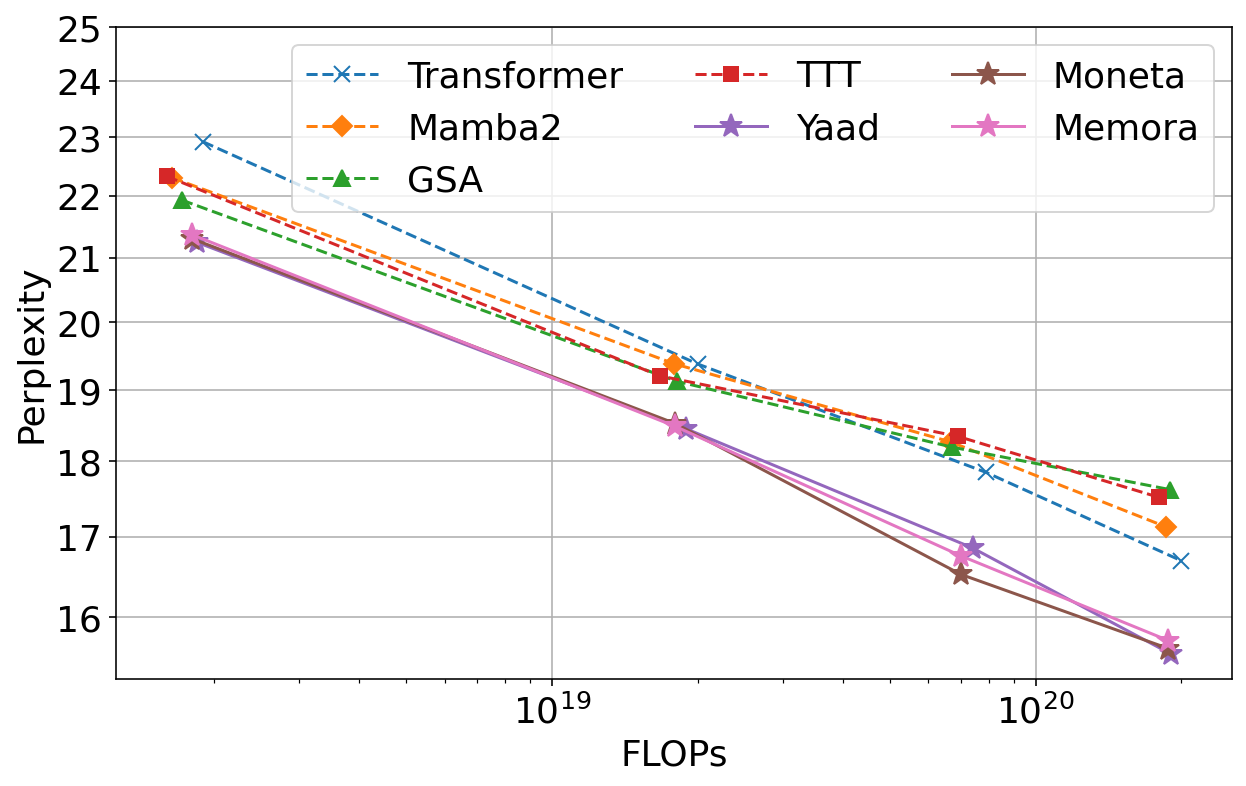}~\hfill{}
    \includegraphics[width=0.32\linewidth]{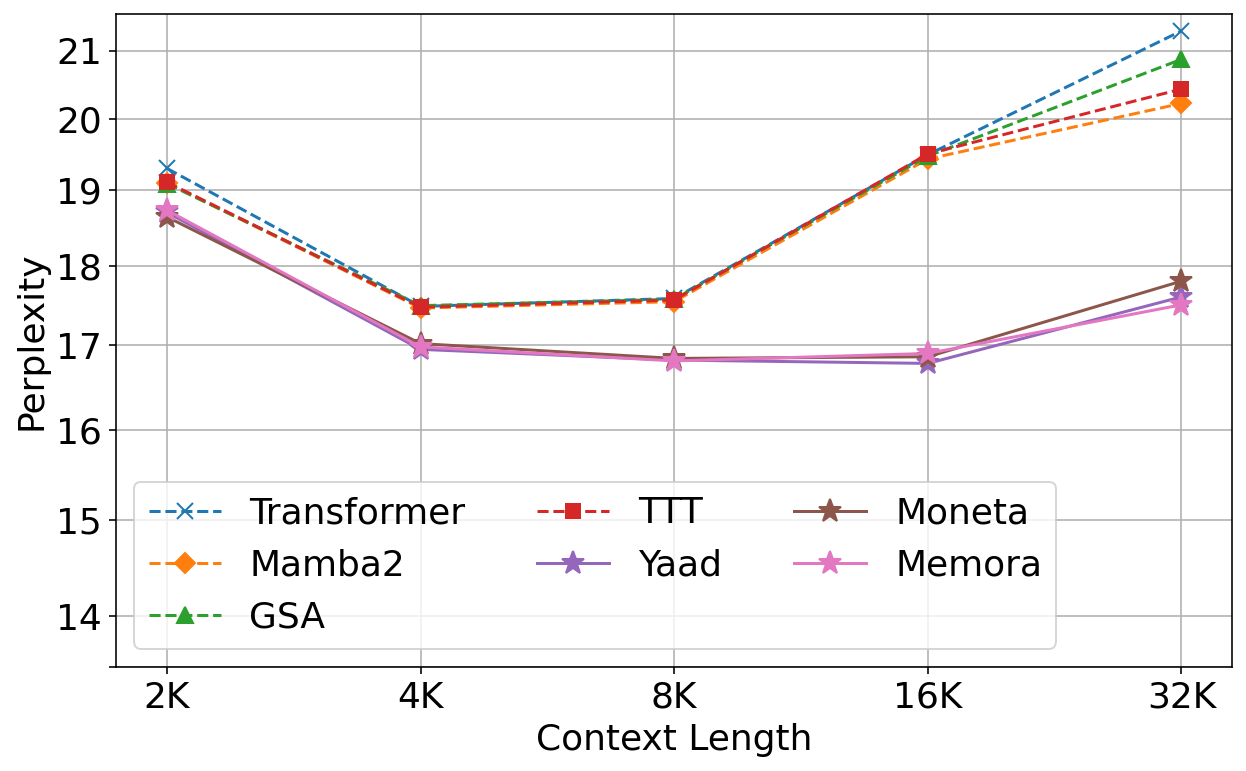}~\hfill{}
    \includegraphics[width=0.32\linewidth]{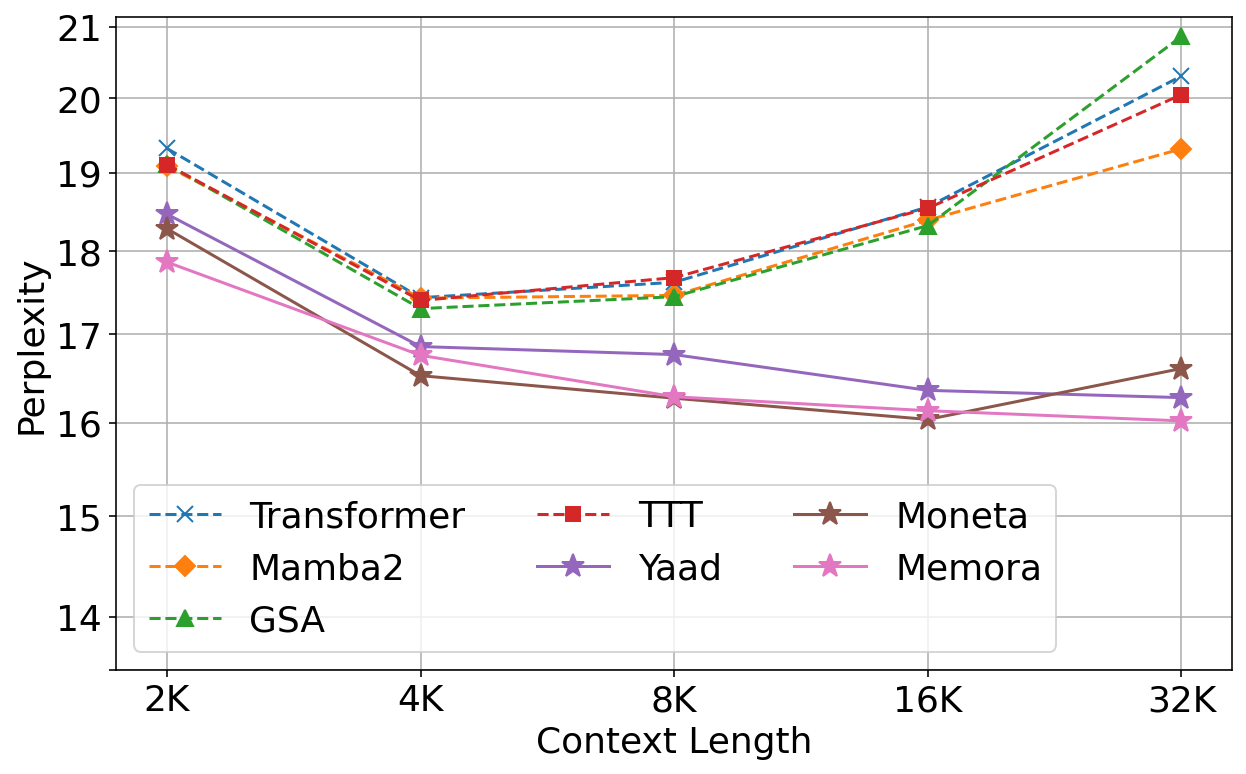}
    \caption{Scaling patterns when increasing (\textbf{Left}) model size, (\textbf{Middle}) sequence length (model size = 340M) (3) (\textbf{Right}) sequence length (model size = 760M) on C4 dataset.}
    \label{fig:scale-pattern}
\end{figure*}

\head{Context Length}
We first vary the training context length from 2K to 32K for two version of our model with size 340M and 760M. The results are reported in \autoref{fig:scale-pattern} (Middle and Right). All three variants of \framework{} scales better than state-of-the-art baselines when increasing the context length. We attribute this superior performance to: (1) expressive memory architecture. Contrary to baselines like Mamba2 and GSA that uses vector- and matrix-valued memory, our variants are using 2-layer MLPs with more expressive power to learn from longer sequences. (2) The choice of retention gate and attentional bias: All of our three variants go beyond the standard attentional biases and retention gates. These choices can help the memory to better manage its fixed-size capacity.

\head{Model Size}
We also report the \#FLOPs vs. perplexity of our models and baselines in \autoref{fig:scale-pattern} (Left). All three variants outperforms all baselines given almost the same budget of FLOPs. These results, once again support the importance of powerful memory design.

\subsection{Needle In Haystack}
To evaluate the effective context window of our models and baselines, we use needle-in-haystack task. In this task, we evaluate the model on retrieving a piece of information (i.e., the ``needle'') from long distractor texts (i.e., the ``haystack''). We focus on the Single NIAH (S-NIAH) task from RULER benchmark~\citep{hsieh2024ruler} and evaluate our models and baselines on sequences with length 1K, 2K, 4K, and 8K. The results are reported in \autoref{tab:haystack}. All our variants outperforms all the baselines with a considerable margin. Interestingly, \mmodel{} shows better performance than others when the data is synthetic noise (S-NIAH-PK). This observation validates the effectiveness of $p$-norm objective and retention gates as they are more robust to noise.  

\begin{table*}
    \centering
    \caption{Performance of \mmodel, \ymodel, \xmodel, and baselines on NIAH task from RULER benchmark. The best results with highest accuracy are highlighted.
    }
    \label{tab:haystack}
    \resizebox{0.65\linewidth}{!}{
    \begin{tabular}{l c c c c c c c c c c}
    \toprule
    \multirow{2}{*}{Model} & \multicolumn{3}{c}{\textbf{S-NIAH-PK}} & \multicolumn{3}{c}{\textbf{S-NIAH-N}} & \multicolumn{3}{c}{\textbf{S-NIAH-W}} & \multirow{2}{*}{\textbf{Average}} \\
    \cmidrule(lr){2-4} \cmidrule(lr){5-7} \cmidrule(lr){8-10}
    &  2K & 4K & 8K &  2K & 4K & 8K  & 1K & 2K & 4K  \\
    \midrule
    \midrule
       Mamba2 & 98.6 & 61.4 & 31.0 & 98.4 & 55.8 & 14.2 & 62.2  & 42.2 & 4.2 & 52.0 \\
       DeltaNet & 96.8 & \cellcolor{myblue}98.8 & 98.6 & 47.2 & 15.4 & 12.8 &  85.2 & 46.2 & 20.0 & 57.9 \\
       Gated DeltaNet & 89.8 & 91.4 & 90.0 & 99.2 & 91.8 & 26.4 & 86.4 & 82.6 & 24.4 & 75.8\\
       TTT  & 98.4 & \cellcolor{myblue}98.8 & 98.0 & 60.2 & 36.6 &  10.2 & 85.8  & 78.8 & 28.0 & 66.1 \\
       \midrule
       \mmodel & \cellcolor{myblue}99.4 & \cellcolor{myblue}98.8 & \cellcolor{myblue}98.8 & 99.4  & \cellcolor{myblue} 99.4  & 92.8 & 92.2 & 88.2 & \cellcolor{myblue} 70.8 &  \cellcolor{myblue}93.5\\
       \ymodel & 99.2 & 98.6 & 94.4 & \cellcolor{myblue} 99.8 & 98.6 & \cellcolor{myblue} 93.2 & 91.8 & \cellcolor{myblue} 89.6  & 67.4 & 92.9\\
       \xmodel & 99.2 & \cellcolor{myblue}98.8 & 92.6 & 98.4 & 99.2 & \cellcolor{myblue} 93.2 &  \cellcolor{myblue} 92.4 & 88.2 & 70.4 & 92.1 \\
    \toprule
    \end{tabular}
    }
\end{table*}

\subsection{Ablation Study}
In this section we perform ablation studies to validate if different design choices that we discussed through the paper are positively contributing for achieving better results.

\head{The Effect of $p$ on Performance}
We first evaluate the effect of $p$ on the performance of \mmodel. We vary the value of $p \in \{1, 1.5, 2, 2.8, 3, 3.2, 4\}$ and context window from 2K to 16K. The results are reported in \autoref{fig:ablation}. Interestingly, there is no monotone pattern when increasing the value of $p$ and the best performance is achieved when $p=3$, while $p = 4$ achieves the worst performance. Also, although different values of $p$ results in different memory modules with varied performance, the scaling pattern when increasing the context length is almost the same.

\head{The Effect of $q$ on Performance}
Similarly, we evaluate the effect of $q$ by varying it in $\{2, 3, 4, 5\}$. Interestingly, contrary to $p$, the value of $q$ can change the scaling pattern when increasing the context length. The main reason for this observation is that the value of $q$ determines the retention gate and a powerful retention gate can improve the memory management, resulting in better performance.

\head{The Effect of Design}
To evaluate the architectural design choices, we perform an ablation study on \ymodel. The results are in \autoref{tab:ablation}. The first row, reports the performance of \ymodel, while (1) the second row removes the retention (i.e., $\beta = 1$), (2) third row make $\delta$ input independent, (3) the third row removes $\ell_2$-loss from the Huber loss, (4) the forth row removes the $\ell_1$ condition, and (5) the last row replaces the MLP with a linear layer. These results indicate that all design choices are contributing to the performance of the model.

\vspace{4ex}

  \begin{minipage}[t!]{\textwidth}
  \begin{minipage}[t]{0.6\textwidth}
    \centering
    \includegraphics[width=0.485\linewidth]{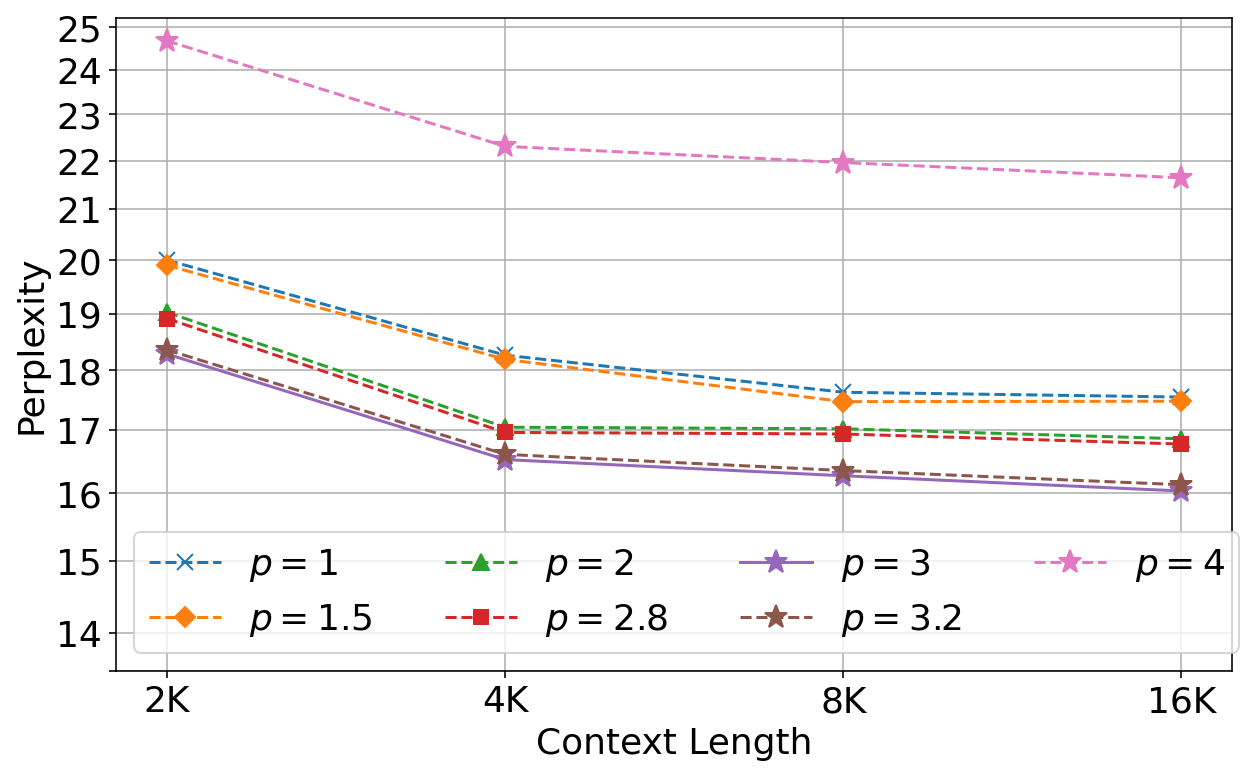} \hfill
    \includegraphics[width=0.485\linewidth]{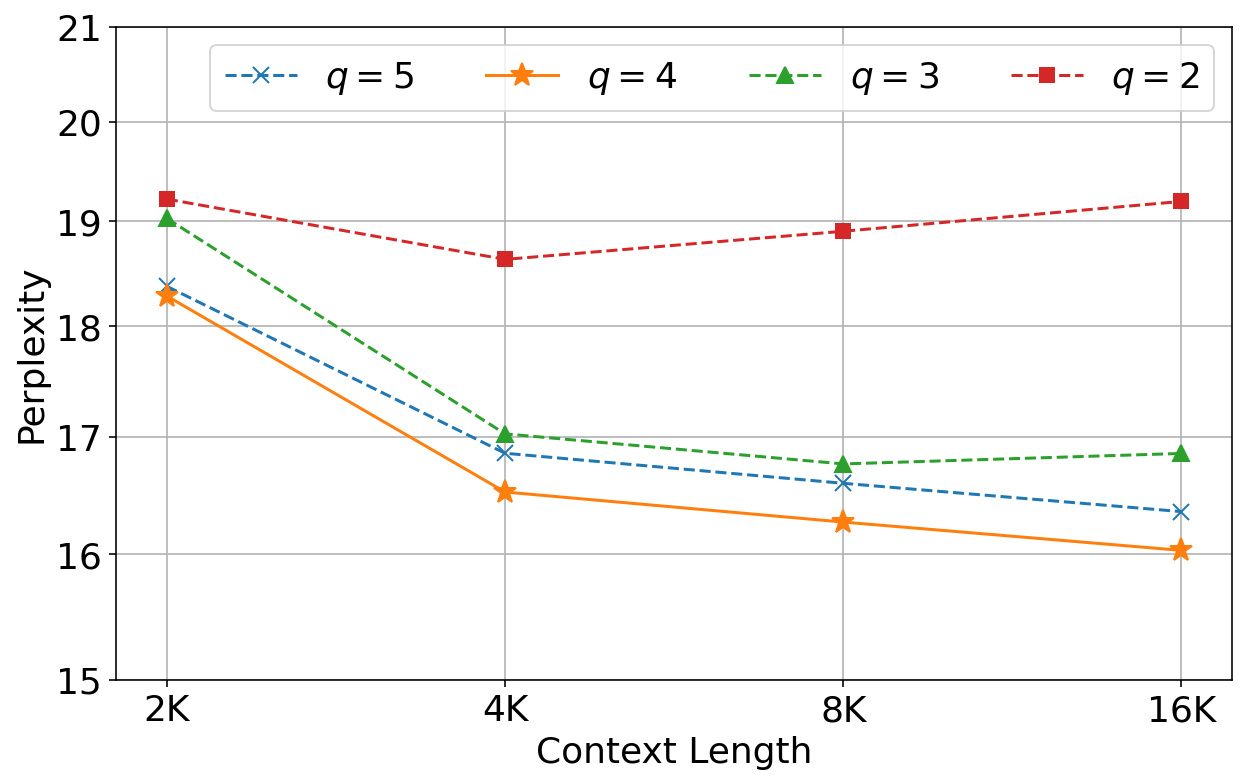}
    \captionof{figure}{The effect of parameters $p$ and $q$ on the performance with different context length.} \label{fig:ablation}
  \end{minipage}
  \hfill
  \begin{minipage}[b]{0.32\textwidth}
    \centering
     \captionof{table}{Ablation study on the components of \ymodel{}.}  \label{tab:ablation}
     \resizebox{0.9\linewidth}{!}{
    \begin{tabular}{lc}
    \toprule
      Model & Avg. LM \\ \hline
        \ymodel &  53.98\\
        \midrule
        - Retention Gate & 50.63 \\
        - Input-dependent $\delta$ & 52.19 \\
        $\ell_2$-loss & 52.86 \\
        $\ell_1$-loss  & 53.04 \\
        linear memory & 51.57\\
    \toprule
      \end{tabular}
      }
    \end{minipage}
  \end{minipage}

\section{Conclusion}\label{sec:conclusion}
In this paper, we present \framework, a general framework that explains the connection of online optimization and test time memorization. \framework{} framework can explain the role of several standard architectural choices in the literature (e.g., forget gate) and helps design next generation of architectures that are capable of managing the memory better. Building upon our framework, we present three novel sequence models, each of which with its own (dis)advantages. Our experimental evaluations show that all these variants are more powerful than Transformers and linear RNNs, in various downstream tasks. In this work, we present a diverse set of variants using \framework. In future, exploring these alternative architectures for different downstream tasks is an interesting future direction.

\newpage
\printbibliography

\newpage
\appendix

\section{Additional Related Work}

\head{Modern Linear RNNs} 
Recent efforts aim to overcome Transformers quadratic cost and limitations in long-context modeling by designing efficient recurrent alternatives~\citep{tiezzi2024resurgence}, mainly due to fast inference and training of such models. The first generation of models–such as RetNet~\citep{sun2023retentive}, LRU~\citep{orvieto2023resurrecting}, RWKV~\citep{peng2023rwkv}, S5~\citep{smith2023simplified}, and S4~\citep{gu2022efficiently}–uses data-independent transition matrix mechanism with Hebbian-like update rule. The second generation of such models started to incorporate input-dependent parameters into such linear architectures (e.g., Griffin~\citep{de2024griffin}, SSMs~\citep{hasani2023liquid, behrouz2024mambamixer, dao2024transformers}, RWKV6~\citep{peng2024eagle}), and/or use more expressive memory updating rule based on delta rule~\citep{peng2025rwkv7, schlag2021linear, yang2024parallelizing, yang2024gated, liu2024longhorn}. The next generation of models, extend the memory architecture to deep models, while using delta-rule-like update rule~\citep{sun2024learning}, or momentum-based update rule~\citep{behrouz2024titans}. Recently, to further enhance the performance of delta-rule-based sequence models, \citet{siems2025deltaproduct} suggest using multiple gradient descent update per token, resulting in more expressive sequence models in state tracking tasks. 

In addition to the above fast linear recurrent sequence models, several studies have focused on (interpretable) non-linear RNNs~\citep{csordas2024recurrent, merrill2024the, lim2024parallelizing,  schone2025implicit, karami2025lattice, von2023uncovering, gonzalez2024towards}, and how their training can be faster~\citep{gonzalez2024towards, lim2024parallelizing, schone2025implicit}. However, due to the recurrent nature of such models, parallelizing them in larger scales is still challenging.

\head{Fast Weight Programs}
The idea of interpretation of linear layers as the key-value associative memory system backs to Hopfield networks~\citep{hopfield1982neural} and then fast weight programs, in which dynamic fast programs are incorporated into recurrent neural networks as writable memory~\citep{schlag2021linear, schmidhuber1992learning, schmidhuber1993reducing}. The two learning rules of Hebbian~\citep{hebb2005organization} and delta rule~\citep{prados1989neural} are the most popular learning rules for them, which have been extensively explored in the literature~\citep{munkhdalai2017neural, schmidhuber1992learning, munkhdalai2019metalearned, schlag2021linear, irie2021going, yang2024parallelizing, yang2024gated}.

\head{Test Time Training}
The key ideas of learning at test time backs to early studies on local learning~\cite{bottou1992local}, in which each test data is trained on its neighbors before making a prediction~\citep{zhang2006svm, gandelsman2022test}. Later applying this idea on modern architectures, it has shown promising performance in diverse downstream tasks such as vision tasks~\citep{jain2011online, mullapudi2019online}, video generation~\citep{dalal2025one}, etc., mostly due to their ability to mitigate out-of-distribution samples.

\head{Hopfield Networks}
We build \framework{} based on the concept of associative memory in its broad form, where we learn an underlying mapping between keys and values. One of the earliest studies that discuss building neural architectures based on associative memory is Hopfield Networks~\citep{hopfield1982neural}, in which associative memory is defined as the minimizing the energy function required to store keys and values. While traditional Hopfield networks has limited applicability in recent years (mainly due to limited capacity of vector-valued memory and energy function), several recent studies aim to improve their capacity by various techniques~\citep{krotov2021hierarchical, li2024expressive, krotov2016dense}, including extending the energy function of such models based on exponential kernels~\citep{krotov2016dense, lucibello2024exponential}, and discuss their connection to Transformers~\citep{ramsauer2021hopfield, hu2024provably}.

\head{Unifying Frameworks}
In recent years, there have been growing efforts to understand the underlying mechanism of sequence models and unify (a subset of) them through a single perspective. \citet{dao2024transformers} present SSD framework to connect linear Transformers and (a subset of) linear recurrent models through the lens of associative operators and structured matrices. The SSD framework, however, is limited to models with vector or matrix-valued memory that are updated using a Hebbian-like update rules. Later, \citet{liu2024longhorn} present an online learning perspective on (a subset of) linear recurrent models. While this framework can also explain more expressive recurrent models based on delta rule, it is limited to online learners (i.e., models that optimize their internal associative memory using stochastic optimizers, such as stochastic gradient descent) with matrix-valued memory. Several modern sequence models, such as Transformers~\citep{transformers} or Titans~\citep{behrouz2024titans} cannot be expressed in this framework.  \citet{sun2024learning} further provide a unifying perspective on how linear and softmax attention are respectively parametric and non-parameteric solutions of (kernel) regression loss but consider other modern linear RNNs outside of this class of models, mainly due to limiting the objective to be regression loss. Recently, in a concurrent work to ours, \citet{wang2025test} also force models to have the same attentional bias objective and show that with additional simplification of modern RNNs (e.g., RetNet~\citep{sun2023retentive}, Mamba~\citep{dao2024transformers}) they approximately place in the same class of models that internally optimize regression loss. However, this simplification, fully change the understanding of underlying update rules in these models. For example, contrary to \citet{wang2025test}, \framework{} can distinguish models with Hebbian-like update (with dot product similarity) and delta rule update (with regression loss). Furthermore, all presented sequence models in this work (e.g., \mmodel, \xmodel, \ymodel) as well as models like HGRN2~\citep{qin2024hgrn} are placed outside of this class of models, due to their different attentional bias.

\section{Proof of Proposition~\ref{prop:Viewpoint1Implies2}} \label{app:proof}
Here we present the proof of Proposition~\ref{prop:Viewpoint1Implies2}. For the sake of completeness, let us first re-state this Proposition.

\noindent\textbf{Proposition~\ref{prop:Viewpoint1Implies2}.} Let $\eta_t = \eta$ and define~$h_t(W) :=  \sum_{i=1}^{t-1} \widehat{\ell}_i (W;\vk_i,\vvv_i) + \frac{1}{\eta}R(W)$. Assume $\cW = \mathbb{R}^d$ and the function $h_t(W)$ is strictly convex in $W$ and let $\mathcal{D}_h(\cdot,\cdot)$ be the Bregman divergence defined by function $h(\cdot)$, i.e., $\mathcal{D}_h(W,W^\prime) = h(W) - h(W^\prime) - \langle \nabla h(W^{\prime}), W-W^{\prime}\rangle$. Set $\mbox{Ret}_t(W,W^\prime) = \mathcal{D}_h(W,W^\prime)$ and $\widetilde{\ell}_t(W;x_t) = \widehat{\ell}_t(W;x_t)$ in \eqref{eq:Viewpoint2-Loss-Premetric}. Then, the update rule in~\eqref{eq:Viewpoint2-Loss-Premetric} is equivalent to the update rule in~\eqref{eq:Viewpoint1-OnlineRegression}. 
\begin{proof}
Let $\{\widehat{W}_1,\widehat{W}_2,\ldots\}$ be the sequence of parameters obtained by \eqref{eq:Viewpoint1-OnlineRegression} and $\{\widetilde{W}_1,\widetilde{W}_2,\ldots\}$ be the sequence of parameters obtained by \eqref{eq:Viewpoint2-Loss-Premetric}. To show both update rules are equivalent, it suffices to show that the above two sequences are the same if they are initialized at the same point. We prove this statement by induction. First of all, since both sequences are initialized at the same point, the induction base is satisfied (i.e. $\widetilde{W}_1 = \widehat{W}_1$. Now, assume by induction hypothesis that 
\begin{equation}
\label{eq:tempProof3}
    \widetilde{W}_{t-1} = \widehat{W}_{t-1}. 
\end{equation}
To complete the induction, we need to show $ \widetilde{W}_{t} = \widehat{W}_{t}$. To this end, notice that,
by \eqref{eq:Viewpoint2-Loss-Premetric}, we have
\[
 \widetilde{W}_{t} = \arg\min_{W} \quad \widetilde{\ell}_t (W,\vk_t,\vvv_t) + \mbox{Ret}_t(W, \widetilde{W}_{t-1})
\]
Using the choice of the Attentional Bias and the Retention function in the Proposition, we obtain
\begin{equation}
\label{eq:tempProof1}
\begin{split}
 \widetilde{W}_{t} = \arg\min_{W} \quad &\widehat{\ell}_t (W,\vk_t,\vvv_t) + \sum_{i=1}^{t-1} \widehat{\ell}_i (W,\vk_i,\vvv_i) + \frac{1}{\eta} R(W) -  \sum_{i=1}^{t-1} \widehat{\ell}_i (\widetilde{W}_{t-1},\vk_i,\vvv_i)\\
    &- \frac{1}{\eta} R(\widetilde{W}_{t-1}) - \Bigg\langle \sum_{i=1}^{t-1} \nabla \widehat{\ell}_i (\widetilde{W}_{t-1}, \vk_i,\vvv_i) + \frac{1}{\eta} \nabla R(\widetilde{W}_{t-1}), W - \widetilde{W}_{t-1} \Bigg\rangle.
\end{split}
\end{equation}
Ignoring the constant terms and using the induction hypothesis~\eqref{eq:tempProof3}, we get 
\begin{equation}
\label{eq:tempProof4}
\begin{split}
 \widetilde{W}_{t} = \arg\min_{W} \quad &\widehat{\ell}_t (W,\vk_t,\vvv_t) + \sum_{i=1}^{t-1} \widehat{\ell}_i (W,\vk_i,\vvv_i) + \frac{1}{\eta} R(W) \\
    & - \Bigg\langle \sum_{i=1}^{t-1} \nabla \widehat{\ell}_i (\widehat{W}_{t-1}, \vk_i,\vvv_i) + \frac{1}{\eta} \nabla R(\widehat{W}_{t-1}), W - \widehat{W}_{t-1} \Bigg\rangle.
\end{split}
\end{equation}
On the other hand, recall that $\{\widehat{W}_1,\widehat{W}_2,\ldots\}$ is obtained by \eqref{eq:Viewpoint1-OnlineRegression}. Therefore, we have
\[
\widehat{W}_{t-1} = \arg\min_{W}\quad \sum_{i=1}^{t-1} \widehat{\ell}_i(W;\vk_i, \vvv_i)  + \frac{1}{\eta} \mathcal{R}_t(W).
\]
Thus,  we have
\begin{equation}
\label{eq:tempProof2}
    \sum_{i=1}^{t-1} \nabla \widehat{\ell}_i (W_{t-1}, \vk_i,\vvv_i) + \frac{1}{\eta} \nabla R(W_{t-1}) = 0.
\end{equation}
Combining \eqref{eq:tempProof2} and \eqref{eq:tempProof4}, we obtain
\[
\widetilde{W}_{t} = \arg\min_{W} \quad  \sum_{i=1}^{t} \widehat{\ell}_i (W,\vk_i,\vvv_i) + \frac{1}{\eta} R(W).
\]
This implies $\widetilde{W}_t = \widehat{W}_t$, which completes the proof.
\end{proof}

\section{Experimental Setup}\label{app:exp-details}
We perform experimental evaluation on the language modeling~\citep{merity2017pointer, paperno-etal-2016-lambada}, common-sense reasoning~\citep{bisk2020piqa, zellers-etal-2019-hellaswag, sakaguchi2021winogrande,clark2018think, clark-etal-2019-boolq}, and long context needle-in-haystack tasks~\citep{hsieh2024ruler}. We compare our models with the state-of-the-art linear recurrent models, Transformers, and hybrid models (recurrent + attention). More specifically we compare with Transformer++~\citep{touvron2023llama}, RetNet~\citep{sun2023retentive}, Gated Linear Attention (GLA)~\citep{yang2024gatedattn}, Mamba~\citep{gu2024mamba}, Mamba2~\citep{dao2024transformers}, DeltaNet~\citep{yang2024parallelizing}, TTT~\citep{sun2024learning}, and Gated DeltaNet~\citep{yang2024gated}.

\begin{table*}[h!]
    \centering
    \caption{Architectural Details.}
    \label{tab:exp-details}
    \resizebox{0.5\linewidth}{!}{
    \begin{tabular}{c c c c c c}
    \toprule
         Model    & Block & Dim & Head & Peak LR & Token\\
         \midrule
         \midrule
        170M & 12 & 768 & 16 & 3e-3 & 15B\\
        340M & 24 & 1024 & 16 & 1.5e-3 & 15B\\
        780M & 24 & 1536 & 16 & 1.25e-3 & 30B\\
    \toprule
    \end{tabular}
    }
\end{table*}



\end{document}